\documentclass{article}

\usepackage{hhline}
\usepackage{arxiv}

\usepackage{url}            
\usepackage{booktabs}       
\usepackage{amsfonts}       
\usepackage{lipsum}
\usepackage{tabularx}
\usepackage{comment}
\usepackage{graphicx}
\usepackage{subfig}
\usepackage{siunitx}
\usepackage{rotating}
\usepackage{pdflscape}
\usepackage{amsmath,amssymb,mathtools}
\usepackage{color}
\usepackage[numbers]{natbib}
\usepackage{csvsimple}
\DeclareMathOperator{\E}{\mathbb{E}}
\title{Convolutional Neural Networks as Summary Statistics for Approximate Bayesian Computation}
\newcommand{\shorttitle}{Convolutional Neural Networks as Summary Statistics for ABC}

\author{
  Mattias~Åkesson\thanks{now at Scaleout Systems AB.}  \textsuperscript{\textdagger},  Prashant~Singh\thanks{Co-First Authors.}, Fredrik~Wrede, Andreas~Hellander \\
  Department of Information Technology\\
  Division of Scientific Computing\\
  Uppsala University\\
  SE 751 05 Uppsala, Sweden \\
  \texttt{andreas.hellander@it.uu.se} \\
}

\begin{document}
\maketitle

\begin{abstract}
Approximate Bayesian Computation is widely used in systems biology for inferring parameters in stochastic gene regulatory network models. Its performance hinges critically on the ability to summarize high-dimensional system responses such as time series into a few informative, low-dimensional summary statistics. The quality of those statistics acutely impacts the accuracy of the inference task. Existing methods to select the best subset out of a pool of candidate statistics do not scale well with large pools of several tens to hundreds of candidate statistics. Since high quality statistics are imperative for good performance, this becomes a serious bottleneck when performing inference on complex and high-dimensional problems.
This paper proposes a convolutional neural network architecture for automatically learning informative summary statistics of temporal responses. We show that the proposed network can effectively circumvent the statistics selection problem of the preprocessing step for ABC inference. The proposed approach is demonstrated on two benchmark problem and one challenging inference problem learning parameters in a high-dimensional stochastic genetic oscillator. We also study the impact of experimental design on network performance by comparing different data richness and data acquisition strategies.  

\end{abstract}

\keywords{Likelihood-free inference \and Summary statistics \and Convolutional neural networks \and Approximate Bayesian computation \and Feature selection}

\section{Introduction}
Likelihood-free parameter inference is a well-studied problem encountered in various domains, most notably including computational biology and astrophysics. The parameter inference problem involves fitting the parameters of a model to observed data from real-world measurements. This allows effective use of simulation models for deeper analysis and understanding of the physical phenomena behind the observed data. The most straight-forward way of estimating parameters in case of tractable likelihood is using Bayesian inference methods and maximum likelihood-estimation. However, for complex models one rarely knows the form of the likelihood functions. For most practical scenarios involving complicated underlying dynamics, likelihood-free parameter inference is the norm. Approximate Bayesian computation (ABC) \cite{beaumont2002approximate,marin2012approximate} has established itself as the most popular likelihood-free inference (LFI) method in the recent past, owing to its flexibility and demonstrated performance on a variety of problems.

Although ABC is a robust LFI method, it involves substantial hyperparameter optimization which makes it challenging to set up optimally \cite{marin2012approximate,sisson2018handbook}, particularly for complex high-dimensional LFI problems involving tens of parameters. The choice of summary statistics is a hyperparameter that presents a great challenge to set up effectively. Summary statistics are typically hand-picked by the practitioner. Automated summary statistic selection methods exist \cite{sisson2018handbook} but these approaches scale poorly as the number of candidate summary statistics increases. Furthermore, optimal summary statistics may not even be present among the initial pool of candidates, which may lead to sub-optimal inference quality. 

Therefore, there has been great interest in developing methods that alleviate cumbersome and explicit summary statistic selection. Kernel embeddings have been explored within the ABC framework to directly compare observed and simulated data by means of a maximum mean discrepancy measure \cite{park2016k2}. \citet{fearnhead2012constructing} show that the best choice for a summary statistic is the posterior mean, when minimizing the quadratic loss. Building upon this theme, recent approaches involve training machine learning (ML) regression models using training data $\mathcal{X}$ that learn the posterior mean $E({\bf \theta} \mid \mathcal{X})$ of parameters ${\bf \theta}$ \cite{10.2307/26384090,2019arXiv190110230W}. The training data $\mathcal{X}$ is composed of pairs $(f({\bf \theta}), {\bf \theta})$, where ${\bf \theta}$ are sampled from a \textit{prior} distribution $p({\bf \theta})$ and $f({\bf \theta})$ corresponds to the realized sample by simulation . The resulting regressor $\hat{{\bf \theta}}(\mathcal{X})$ captures various characteristics of $\mathcal{X}$ and can be used as a summary statistic within the ABC framework. 

This paper makes two key contributions to the emerging class of LFI methods using artificial neural networks as the regression model \cite{Cranmer2020}. First, we propose a convolutional neural network (CNN) architecture that learns the estimated posterior mean $\hat{{\bf \theta}}(\mathcal{X})$. The CNN is particularly effective towards learning local features that can distinctly characterize various intricate patterns within time series responses. Our intuition is that this will lead to more accurate modeling of the posterior mean, and in turn enhances inference quality. 
The proposed CNN architecture is specifically tailored for inference on high-dimensional stochastic biochemical networks models realized by continuous-time discrete-space Markov chains which lack tractable likelihoods. These models have inherent intrinsic noise associated with low copy-numbers of chemical species. Furthermore, for a typical gene regulation network model a dimensionality of the reaction rate parameter vector below 10 is considered to be a small model. Thus, parameter inference of realistic biochemical networks is often very challenging. To our knowledge, previous implementations of neural networks for LFI have only considered fairly low-dimensional test problems \cite{10.2307/26384090,2019arXiv190110230W}. Our second key contribution is thus a systematic evaluation of our new CNN approach as well as two previously suggested architectures on a realistic, high-dimensional stochastic inference problem in computational systems biology. The experiments conducted in this work demonstrate that the artificial neural network (ANN) approach to LFI shows promise towards tackling real-world problems, with the CNNs emerging as a particularly scalable solution.       

Section \ref{sec:background} formally introduces the likelihood-free parameter inference problem, and briefly describes existing methods, including ABC and ANN based methods. Section \ref{sec:method} briefly explains CNNs and presents the proposed CNN architecture for learning summary statistics. Section \ref{sec:experiments} describes the experimental settings and Section \ref{sec:res} demonstrates the performance of the proposed approach on test problems, and compares the results with the state of the art. 
Section \ref{sec:conclusion} concludes the paper.


\section{Background and Related Work}
\label{sec:background}
Consider an observed dataset $X$ and a simulator or analytical model $f({\bf \theta})$ corresponding to the physical process that generated $X$. The parameter inference task in a likelihood-free setting is to infer the value of parameters ${\bf \theta}$ that results in simulator output $f({\bf \theta})$ agreeing with observed data $X$.
As inference in a likelihood-free setting must proceed solely using access to the simulator $f({\bf \theta})$ and observed dataset $X$, sampling candidates ${\bf \theta}$ and comparing simulated responses to $X$ forms the basis of ABC. The ABC \textit{rejection sampling} algorithm begins by sampling candidates ${\bf \theta} \sim p({\bf \theta})$, where $p({\bf \theta})$ is the \textit{prior} distribution encoding prior knowledge about the problem. The sampled ${\bf \theta}$ is then simulated and the response ${\bf y} = f({\bf \theta})$ is compared to $X$. Since simulation outputs are typically high-dimensional (e.g., time series), the comparisons are instead made in terms of low-dimensional features or \textit{summary statistics} ${\bf S} = \{S_1({\bf y}), ..., S_n({\bf y})\}$. The simulated response ${\bf y}$ can then be compared to $X$ using a distance function $d$ as $d_{sim} = d({\bf S}({\bf y}), {\bf S}(X))$. Given a tolerance threshold $\tau$, if $d_{sim} \leq \tau$ the corresponding ${\bf \theta}$ is deemed to be accepted, or otherwise rejected. This rejection sampling cycle proceeds until a specified number of samples have been accepted, forming an empirical estimation of the posterior distribution $p({\bf \theta}|X)$.

The motivation for the use of expressive low-dimensional summary statistics ${\bf S}$ is due to the \emph{curse of dimensionality}. When ${\bf S}$ is high-dimensional (or when no summary statistics are used), there is a greater likelihood of random discrepancies between ${\bf S}({\bf y})$ and ${\bf S}(X)$ \cite{prangle2015summary}. For high-dimensional summaries, a larger threshold may be required in order to achieve a reasonable number of accepted samples. Consequently, the quality of approximation of the posterior will suffer in such a case. 

As the summary statistics form the basis of the comparison between simulated responses and observed data, the choice and subsequent quality of used statistics is paramount towards achieving high quality inference. Substantial effort has been invested in research towards summary statistic selection \cite{sisson2018handbook}[Chap. 5]. 
However, to circumvent the problem of selecting sub-optimal summary statistics, recent advances in automating summary statistic learning using regression models are of particular interest.

\subsection{Estimated posterior mean as a summary statistic}
\citet{fearnhead2012constructing} presented a regression-based approach towards constructing summary statistics where, for ${\theta}_j, j=1,...,L$, a linear regression model of the form,
\begin{align}
    {\theta}_j^i &= E({\theta}_j \mid {\bf y}^i) + \sigma_j \xi^i, \text{where,}\\
    E({\theta}_j \mid {\bf y}^i) &= b_{0,j} + b_j h({\bf y}^i)
    \label{eq:linear_reg}
\end{align}
with ${\bf y}_i$ being the $i$-th simulated sample or observed data sample, $h$ the vector-valued transformation function, $\sigma_j$ the scale parameter of the $j$-th linear regression, and $\xi^i$ is Gaussian mean-zero noise. The model parameters in (\ref{eq:linear_reg}) are fitted using least-squares on a simulated dataset $D = \{{\theta^i, {\bf y}^i}\}^N$, where ${\bf \theta}^i \sim p({\bf \theta})$. The estimated mean posterior represented by the $L$ linear regression models can then be used as a summary statistic ${\bf S}$ within ABC rejection sampling. The dataset $D$ makes use of $p({\bf \theta})$ but is distinct from the simulations used for rejection sampling. Therefore, the statistic selection process entails significant overhead.

The training set $D$ is used to fit model parameters of the regressor, but it can also be used for data efficiency to perform ABC in a \textit{reference table} scenario \cite{cornuet2008inferring}. The reference table method entails computation of distance values $d_{sim}^i = \{d({\bf S}({\bf y}^i), {\bf S}(X))\}_{i=1}^N$. 
The samples comprising the smallest $x$-th percentile of all distances are deemed to be accepted samples and form the ABC estimated posterior. The reference table method allows for reusing training data in subsequent ABC rejection sampling, enabling better data efficiency. The ABC reference table method is used in the ANN based methods described below, as well as in this work.

As an alternative non-linear approach, a deep neural networks was proposed by \citet{10.2307/26384090} to estimate the posterior mean, in hope of learning even more informative summary statistics as opposed to linear regression. The  dense (deep) neural network (DNN) model is the simplest ANN model, it consists of multiple layers of interconnected neurons. 
The DNN based summary statistic construction in \cite{10.2307/26384090} was shown to outperform the linear regression method, though at additional computational cost as the DNN requires more training data.

A novel ANN architecture named partially exchangeable networks (PEN) was proposed by \citet{2019arXiv190110230W}. The model is a generalization of the Deep Sets model, an ANN model using sets instead of ordered data as input. The PEN model extends the idea of sets for data with $d$-partially exchangeable structures in a conditionally Markovian context. 
The authors show results for 4 different stochastic models, two of which involve time series data: a stochastic auto-regressive time series model of order 2 and the Moving Average of order 2 also used in \cite{10.2307/26384090}. The results shows that the PEN models produce a more reliable posterior even when using less training data compared to the DNN. 


Although the PEN architecture reduces the number of trainable weights of the ANN (and in turn increases ANN model efficiency) by leveraging partial exchangeability, we believe there is room to improve the expressive power of the ANN model by exploiting rich local patterns present within temporal responses. We propose a general convolutional neural network (CNN) architecture wherein a sequence of convolutional layers extracts specific local patterns within the input time series. These rich local patterns allow the CNN model to incorporate effective discriminative abilities for input patterns, that are critical in an informative summary statistic. The aim of this work is therefore to develop a CNN architecture that exceeds current state of the art ANN summary statistic models in terms of informativeness and subsequent ABC inference quality for complex large-scale problems, while being data-efficient.
Embedded neural networks have also been used in non-ABC LFI settings to learn summary statistics from simulated data. In such settings, neural density estimators \cite{lueckmann2021benchmarking} make use of the embedded networks to either estimate the posterior directly (e.g., combining a summary statistic network with an inference network \cite{radev2020bayesflow}), or use synthesized likelihoods which requires Markov chain Monte Carlo (MCMC) sampling \cite{Cranmer_frontierSBI}. In this work we focus solely on obtaining expressive summary statistics using ANN architectures that can be used in LFI settings (i.e., both the ABC-family of methods such as ABC-Sequential Monte Carlo and ABC-Markov Chain Monte Carlo and non-ABC LFI methods such as those based on density estimation).
The following section explores our proposed CNN architecture in detail.

\section{Convolutional Neural Networks}
\label{sec:method}
The inherent structure in time series makes convolutional networks an attractive option to explore for the task of learning the mapping between time series responses as input to the CNN, and the posterior mean $\hat{\theta}$ as output of the CNN. The CNN will effectively incorporate summary statistics in its hidden layers and can subsequently be used in conjunction with existing likelihood-free inference methods for parameter inference, or to perform model exploration where the goal is to screen the parameter space for different qualitative behaviors produced by the model \cite{Wredesmart2019}.

CNNs form an architecture of neural networks for processing data having a grid-based structure. Temporal data in the form of time series is often obtained at regular intervals, forming a 1-dimensional grid structure. This is certainly true for time series data originating from simulations where it is possible to have time series values at specific time points. This property makes CNNs particularly suited for estimating the posterior mean and the input patterns are time series sampled at regular intervals. 

A CNN replaces general matrix multiplication in a multi-layer neural network with the convolution operation in at least one of the layers. The convolution operation enables performing weighted averaging of inputs such that more recent entities in the input are given larger weights. Intuitively, this allows for identification of local informative patterns in data. For example, in case of time series as input, the convolution operation can be used to identify distinct behaviors such as maxima, distance to first peak, etc. No hand-crafting of features is necessary. Formally, for input data $y$ and a kernel $w$, the discrete convolution operator can be defined as follows \cite{goodfellow2016deep},
\begin{equation}
    s(t) = (y * w)(t) = \sum_{a=-\infty}^{\infty} y(a)w(t-a),
\end{equation}
where $t$ is a specific time point. The kernel $w$ is essentially a filter represented by a matrix of trainable weights. The kernel matrix is typically small and is applied to a small region of the input. In practice, a fixed finite filter window is used (e.g., of size $3$ as is used in this work). By operating as a filter, the kernel is able to enable detection of features such as edges of objects within an image. In case of time series, such features would include various characteristics of the time series such as distinct types of peaks.

\begin{figure*}
\centering
  \begin{tabular}{@{}c@{}}
    \includegraphics[width=\textwidth]{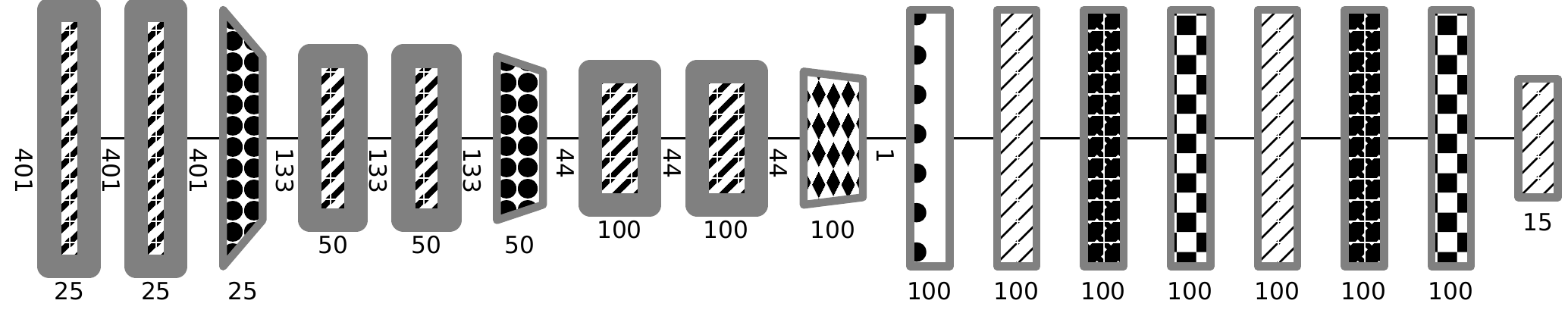}\\
    \includegraphics[width=\textwidth]{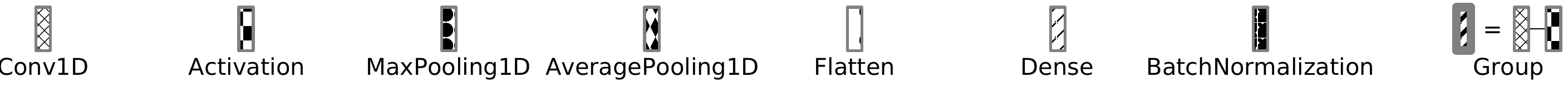}
  \end{tabular}
  \caption{A schematic view of the convolutional neural network architecture used in this work for the genetic oscillator test problem - $(401 \times 1)$ time series as input, $15$ predicted parameters as output. The numbers at the bottom of each layer denote the layer output space dimensionality (number of convolutional filters). The convolutional window size is set to 3.} Visualized using Net2Vis (https://github.com/viscom-ulm/Net2Vis).\label{graph}
\end{figure*}

Figure \ref{graph} depicts the CNN architecture used for experiments in this article concerning the genetic oscillator test problem described later. The input layer of dimensionality $(401 \times 1)$ accepts the time series input, where each time series is composed of $401$ values. A sequence of convolutional and pooling layers then operate on the time series where the convolution operator identifies local patterns in the input to the layer, and subsequently the pooling operation replaces the output at certain places with a feature of nearby outputs. Specifically, we use max pool \cite{zhou1988computation} where the maximum value of the output within a rectangular neighborhood is chosen \cite{goodfellow2016deep}. The pooling layer thus achieves dimensionality reduction or in essence, feature selection from the convolution layer where it receives input from. The effect of pooling is also that the size of the network decreases, reducing the computational complexity.
After $2$ combinations of convolution and max pooling, the output is processed through a layer of average pooling and subsequently through $2$ dense layers before finally reaching the output layer representing the estimated posterior mean.

\section{Experimental Setup}
\label{sec:experiments}
The experiments are designed to evaluate the informativeness of the CNN-based summary statistic in the context of ABC parameter inference. 
The proposed CNN architecture is evaluated and compared to the DNN \cite{10.2307/26384090} and PEN \cite{2019arXiv190110230W} architectures. The term ANN is used henceforth to refer to either of the DNN, PEN and CNN architectures.
The likelihood-free parameter inference pipeline using the ANN-based summary statistics consists of the following steps.
\begin{enumerate}
    \item Generate training data for the ANN: draw $N$ samples from a uniform prior defined over a specified range, and simulate the corresponding time series. 
    \item The ANN regression model is trained on the $N$ samples above, and is used to predict the posterior mean for some observed data.
    \item ABC inference: the predicted posterior mean is used as a summary statistic within the framework of ABC rejection sampling. The reference table method described earlier is used in the experiments, and utilizes pre-generated data distinct from training data.
\end{enumerate}
All experiments have been conducted using the freely available scalable inference, optimization and parameter exploration (sciope) Python3 toolbox \cite{sciope}. Sciope implements all 3 ANN architectures considered in this work.

The following text describes the experimental setup with respect to quantifying the summary statistic posterior estimation error, and the ANN model training framework.

\subsection{Summary Statistic Posterior Estimation Error}
In order to evaluate the goodness of ANN-based summary statistics, the quantity of expected distance can be denoted as follows,
\begin{equation}
\E_{\theta, y} [d(\hat{\theta}(y),\theta)],
\label{expected_dist}
\end{equation}
where $\hat{\theta}$ is the posterior mean estimated by the ANN model, $\theta \in p(\theta)$ represents the true parameter values and $d$ is a given distance function. The choice of $d$ in this work is the absolute value of the Euclidean distance, while $p(\theta)$ is the uniform prior.
The expected distance as defined in Eq. (\ref{expected_dist}) is intended to measure the error in estimation of the posterior mean when using an ANN-based summary statistic.

A measure independent of the considered prior range is desirable. The normalized mean absolute error (MAE) is defined as,
\begin{equation}
    E_{\%} = \frac{\E_{\theta, y} [d(\hat{\theta}(y),\theta)]}{d({\theta}_m,\theta)},
    \label{info_gain}
\end{equation}

where the denominator is the MAE based on the prior knowledge, e.g., the prior mean $\theta_m$. This allows capturing the new information gained by the regression-based ANN models over the prior knowledge. $E_{\%} = 1$ indicates no new information gained while $E_{\%} < 1$ indicates relative accuracy improvements or new information gained by the regression model. 
A uniform prior $U(\bf{dmin}, \bf{dmax})$ is used, resulting in the denominator taking the form,

\begin{equation}
    d({\theta}_m,\theta)  =  \frac{\bold{dmax} - \bold{dmin}}{ 4}.
\end{equation}
\noindent
The numerator can be approximated using a set of $n$ test points as,

\begin{equation}
     \E_{\theta, y} [d(\hat{\theta}(y),\theta)] \approx \frac{1}{n} \sum_{i=1}^n |\theta_i - \hat{\theta}(y_i)|.
\end{equation}

Equation (\ref{info_gain}) can now be rewritten as,
\begin{equation}
    E_{\%} \approx \frac{4 }{ \bold{dmax} - \bold{dmin} } \frac{1}{n} \sum_{i=1}^n |\theta_i - \hat{\theta}(y_i)|.
\end{equation}
Note that $E_\%$ is not solely a function of the accuracy of the ANN, it is also related to the numerical, or practical, identifiability of the given model and parameters. Depending on the observed data, a substantial information gain, i.e., an $E_\% << 1$ might not be observed using any available inference method. However, for those parameters we can identify, $E\%$ provides an effective means of comparing the different ANN architectures. For this reason, we conduct numerical experiments where we also vary the observed output state variables and the amount of observed data, in addition to the amount of samples from the prior used to train the ANNs.

\subsection{Model Training}
\label{sec:model_training}
The training data corresponding to the DNN, PEN and CNN models is pre-computed and is the same for all three architectures for a given experiment. For each layer type, the layer count and width have been kept consistent across architectures in order to minimize the effect of architecture depth and scale. For example, the $\text{PEN}_{10}$ architecture used in experiments corresponding to the CNN architecture shown in Figure \ref{graph} has the same number and scale of convolutional and dense layers as the CNN. The corresponding DNN has the same number and scale of dense layers as the CNN and $\text{PEN}_{10}$.

Two model training approaches have been explored. A single-shot approach (\textit{approach 1}) involving gradient descent using a batch size of 512, and a two stage approach (\textit{approach 2}) involving two different batch sizes of training data. In the first stage, a relatively small batch size of 32 is used and stochastic gradient descent is used to optimize the ANN model hyperparameters. The numbers of training epochs is determined by the early stopping regularization with the patience parameter being 5 epochs (as in the first approach as well). In the second stage, a batch size of 4096 is used, along with the same early stopping criterion described above. The motivation for starting with a smaller batch size is as follows.

The ratio of learning rate to the batch size in an important factor controlling stochastic gradient descent (SGD) dynamics \cite{jastrzkebski2017three}. In particular, for a majority of the the SGD training update steps, the search moves between valley-like regions of the loss function landscape at a height above the valley floor \cite{xing2018walk}. This phenomenon allows the optimization process to initially cover greater distance towards the optima as compared to starting with a larger batch size. Consequently, the search towards the optima is accelerated allowing for better generalization for a fixed number of epochs. For a deeper discussion on the effect of batch sizes and learning rates, the reader is referred to \cite{jastrzkebski2017three,xing2018walk}.

The loss function for model training is the mean squared error (MSE) on the training set, while the early stopping criterion involves calculation of the mean absolute error (MAE) using the validation set. The test set is finally used to calculate the expected estimation distance as in Eq. (\ref{expected_dist}). \textit{Approach 2} entails substantially longer training time but delivers quantifiable improvements in model accuracy for all three architectures (e.g., Tables \ref{tab:lv_ds}, \ref{tab:ds_sizes}). \textit{Approach 1} is used for all except two experiments where it is explicitly mentioned.

\section{Results}
\label{sec:res}
The proposed approach is demonstrated on three test problems. The Lotka-Volterra predator-prey model and the moving average 2 (MA2) model are benchmark parameter inference test problems in literature, and serve to effectively compare the proposed approach to existing methods. The genetic oscillator is a challenging high dimensional test problem and serves to demonstrate the scalability of the proposed approach.

\begin{table}
\small
\centering
\scalebox{1.0}{%
\setlength{\tabcolsep}{10pt}
\begin{tabular}{c|c|c}
Network & $E_{\%}$ & $E{\%}_{true}$\\
\hline
& \multicolumn{2}{c}{ Training Set Size $10^3$}\\ 
\cline{2-3}
$DNN$ & $0.676 \pm 0.037$ & $0.464 \pm 0.282$\\
$PEN_{10}$ & ${\bf 0.499 \pm 0.133}$ & ${\bf 0.326 \pm 0.172}$\\
$CNN$ & $0.531 \pm 0.180$ & $0.376 \pm 0.086$\\
& \multicolumn{2}{c}{ Training Set Size $10^4$}\\
\cline{2-3}
$DNN$ & $0.474 \pm 0.010$ & $0.431 \pm 0.262$\\
$PEN_{10}$ & $0.221 \pm 0.005$& ${\bf 0.173 \pm 0.102}$\\
$CNN$ & ${\bf 0.198 \pm 0.007}$& $0.189 \pm 0.095$\\
& \multicolumn{2}{c}{ Training Set Size $10^5$}\\
\cline{2-3}
$DNN$ & $0.197 \pm 0.004$ & $0.200 \pm 0.124$\\
$PEN_{10}$ & ${\bf 0.159 \pm 0.003}$ & ${\bf 0.154 \pm 0.102}$\\
$CNN$ & $0.183 \pm 0.009$ & $0.190 \pm 0.122$\\
& \multicolumn{2}{c}{ Training Set Size $10^6$}\\
\cline{2-3}
$DNN$ & $0.158 \pm 0.002$ & $0.168 \pm 0.095$\\
$PEN_{10}$ & ${\bf 0.149 \pm 0.002}$ & ${\bf 0.154 \pm 0.082}$\\
$CNN$ & $0.165 \pm 0.010$ & $0.176 \pm 0.094$\\
\end{tabular}}
\caption{$E_{\%}$ for inference on the MA(2) model for training set sizes $10^3 - 10^6$. $E{\%}_{true}$ is calculated in relation to $\theta_{\text{true}} = (0.6, 0.2)$ instead of a uniformly generated test set, and is often used \cite{10.2307/26384090,2019arXiv190110230W} in benchmark parameter inference experiments with the MA(2) model.} The values represent the mean and standard deviation over 10 independent experiments.
\label{tab:ma2}
\end{table}

\subsection{The Moving Average 2 Model}
\label{sec:ma}
The moving average model is a relatively simple and  popular benchmark example used in ABC \cite{marin2012approximate} and ANN summary statistics literature \cite{10.2307/26384090,2019arXiv190110230W}. The typical model setting considered herein (and in works above) allows exact calculation of the posterior distribution. Manually selected summary statistics for the moving average model include autocovariance at various lag intervals, and have been extensively studied \cite{marin2012approximate,2019arXiv190110230W}. The moving average model is therefore a good choice for benchmarking new summary statistic selection methods in an ABC context. The experimental settings follow \cite{2019arXiv190110230W}.

The moving average model of order $q$, MA($q$) is defined for observations $X_1, ..., X_p$ as \cite{10.2307/26384090},
\begin{equation*}
    X_j = Z_j + \theta_1 Z_{j-1} + \theta_2 Z_{j-2} + ... + \theta_q Z_{j-q}, j=1,...,p,
\end{equation*}
where $Z_j$ represents latent white noise error terms. This work considers $q=2$ with experimental settings matching \cite{10.2307/26384090,2019arXiv190110230W} including $Z_j \sim N(0,1)$. The MA(2) model is identifiable in the following triangular region,
\begin{equation*}
    \theta_1 \in [-2,2], \theta_2 \in [-1, 1], \theta_2 \pm \theta_1 \geq -1.
\end{equation*}
The training data for all ANN architectures is sampled uniformly over this region. The training, validation and test set sizes are set to $10^6, 10^5, 10^5$ samples respectively, matching the configuration in \cite{10.2307/26384090}. The DNN architecture (3-layer, 100 neurons per layer) is also set to mirror the settings in \cite{10.2307/26384090}. The evolution of ANN model accuracy with varying size of training data is also explored, in addition to overall model accuracy over $10^6$ training samples. Model training \textit{approach 1} (Sec \ref{sec:model_training}) is used in all experiments for this test problem.

Table \ref{tab:ma2} compares the performance of the DNN, PEN and CNN architectures on the MA(2) model. The configuration for the $\text{PEN}_{10}$ variant follows \cite{2019arXiv190110230W}. 
The performance of all architectures is comparable for the relatively simple MA(2) model. It can be observed that the PEN and CNN architectures outperform DNN in a majority of cases, especially for smaller training sets. As the training sets grow in size, the performance deficit between the architectures diminishes substantially.

\begin{figure*}
\subfloat[DNN - training set of $10^3$ samples.]{\label{fig:figure14_1}\includegraphics[width=.3\linewidth]{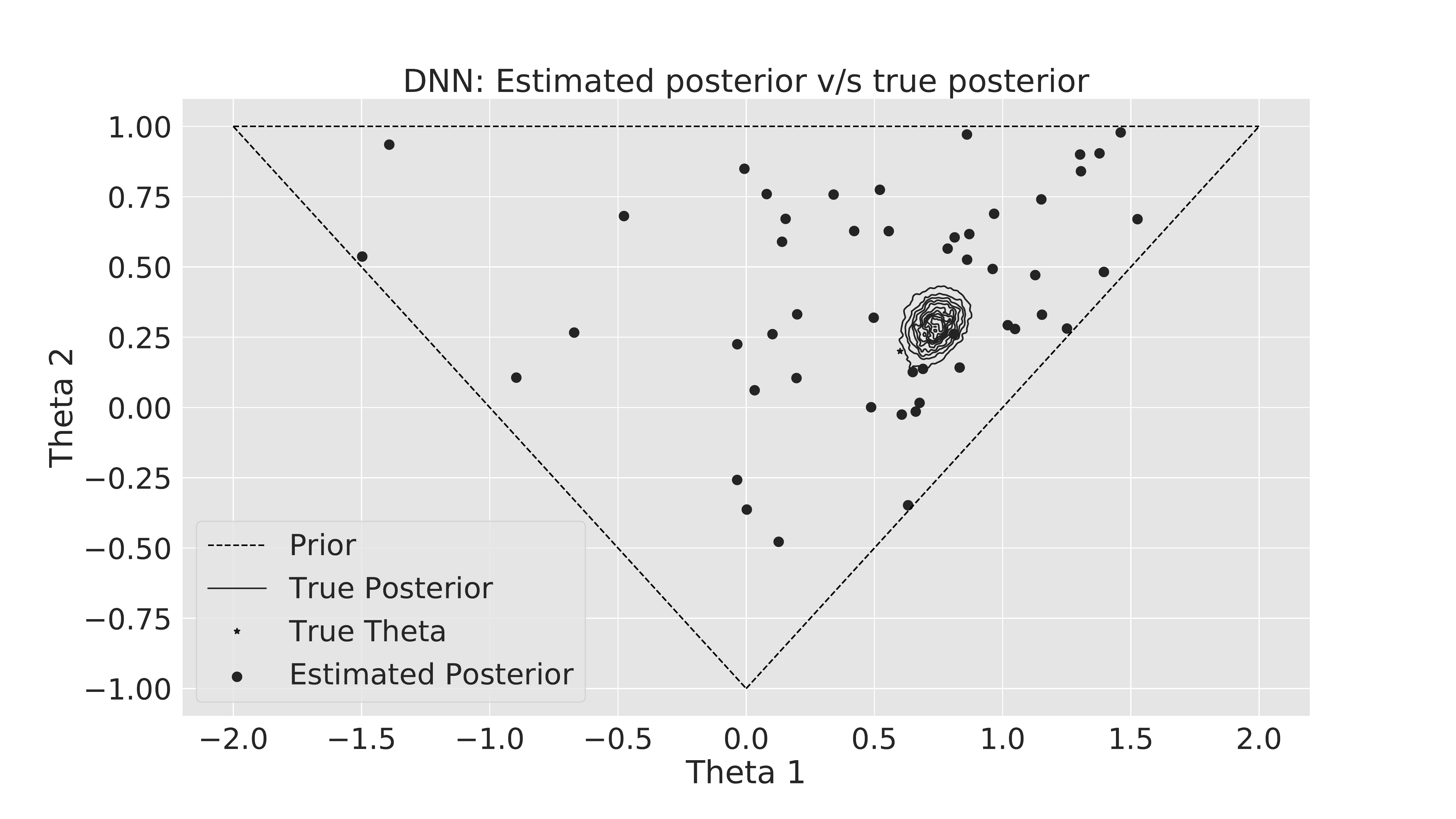}}
\hfill
\subfloat[PEN - training set of $10^3$ samples.]{\label{fig:figure14_2}\includegraphics[width=.3\linewidth]{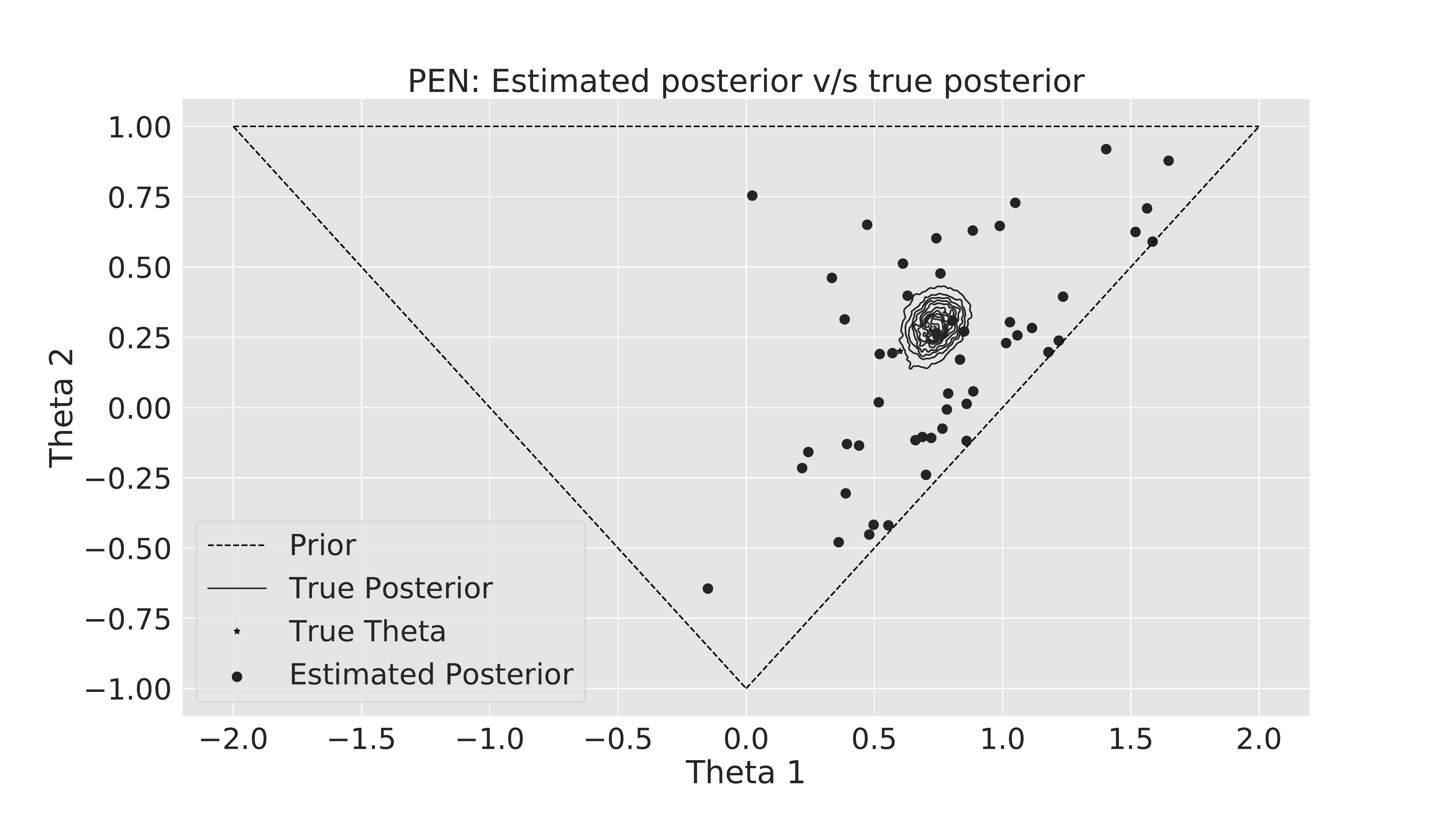}}
\hfill
\subfloat[CNN - training set of $10^3$ samples.]{\label{fig:figure14_3}\includegraphics[width=.3\linewidth]{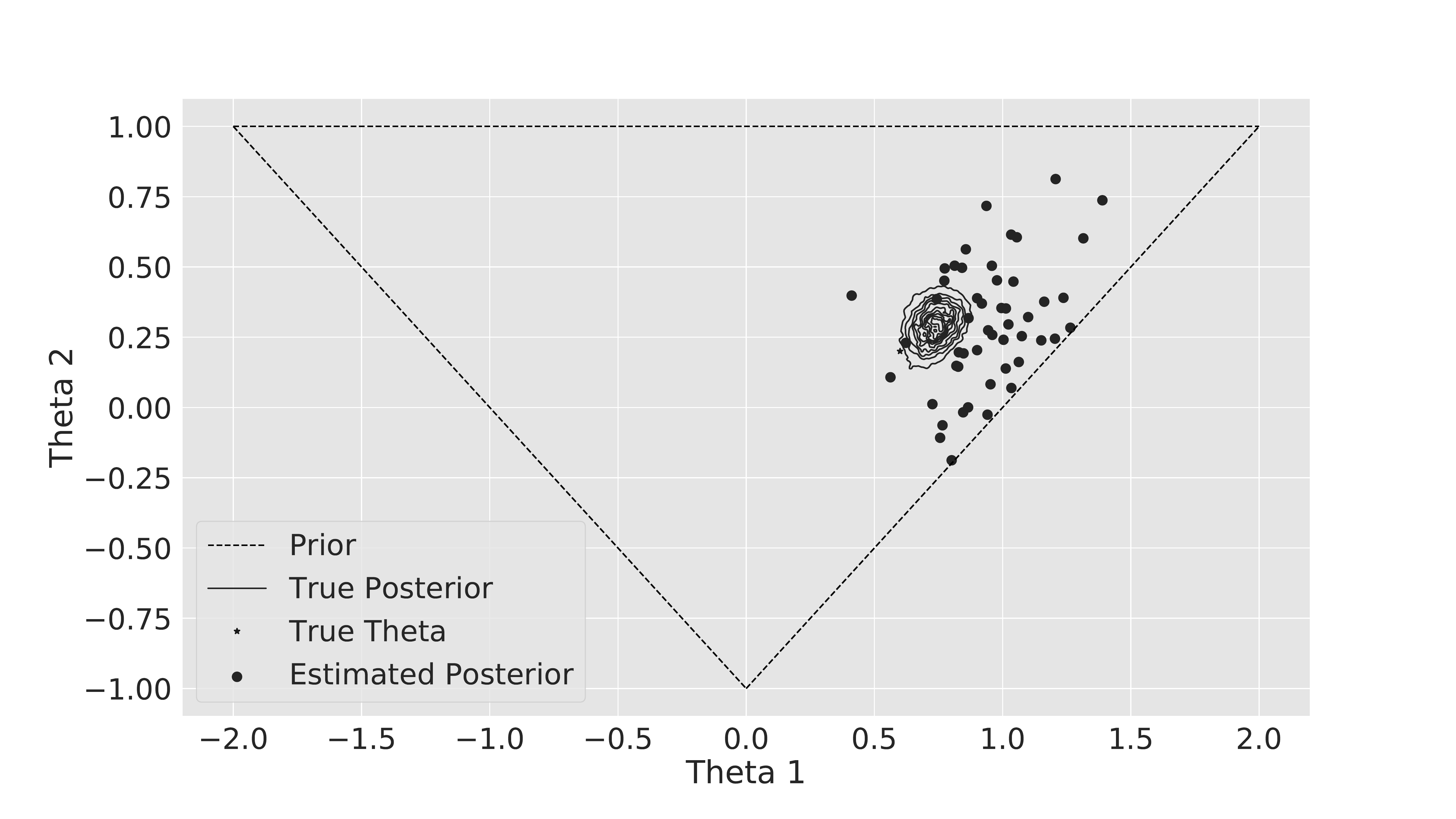}}

\subfloat[DNN - training set of $10^4$ samples.]{\label{fig:figure14_4}\includegraphics[width=.3\linewidth]{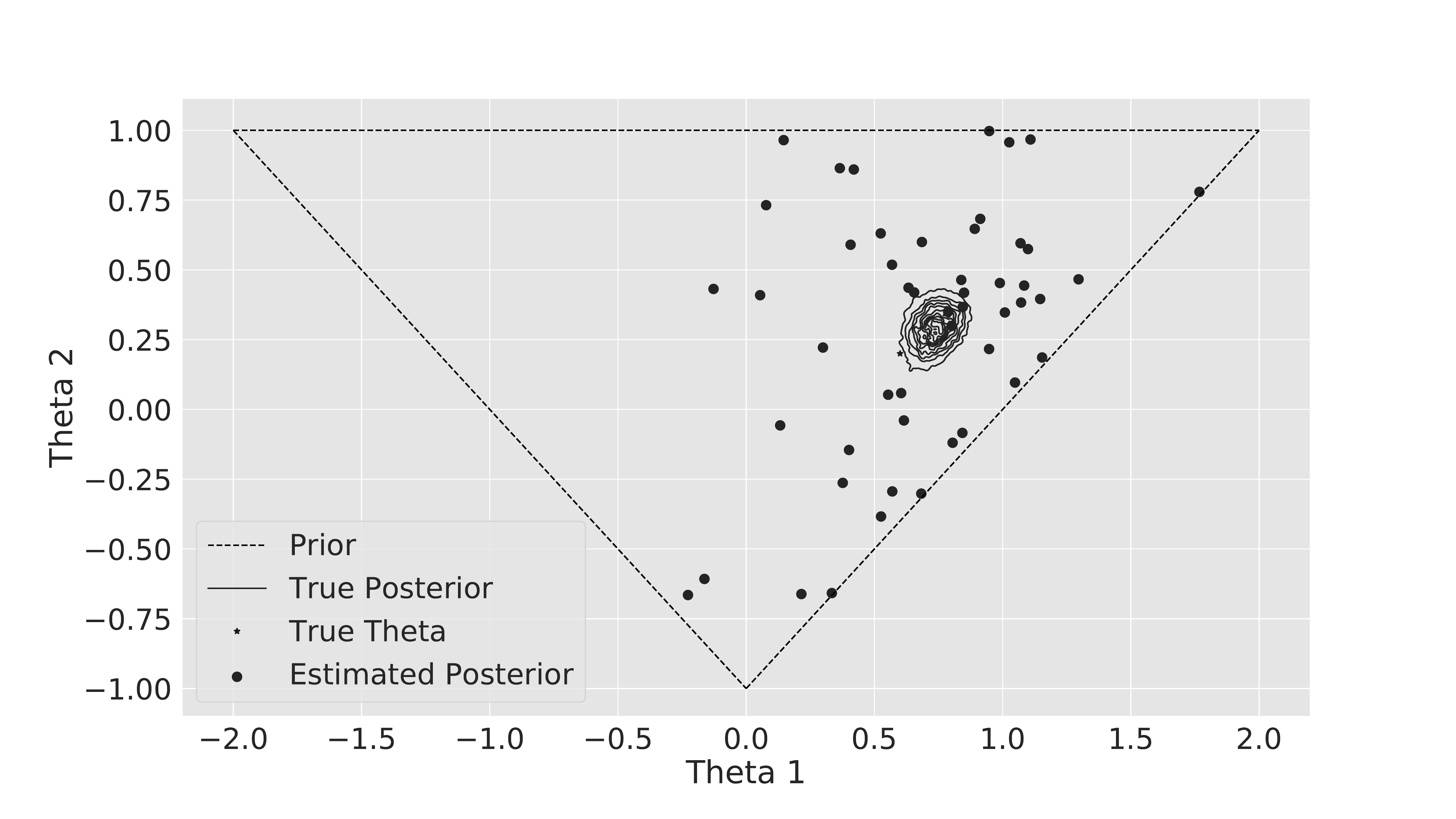}}
\hfill
\subfloat[PEN - training set of $10^4$ samples.]{\label{fig:figure14_5}\includegraphics[width=.3\linewidth]{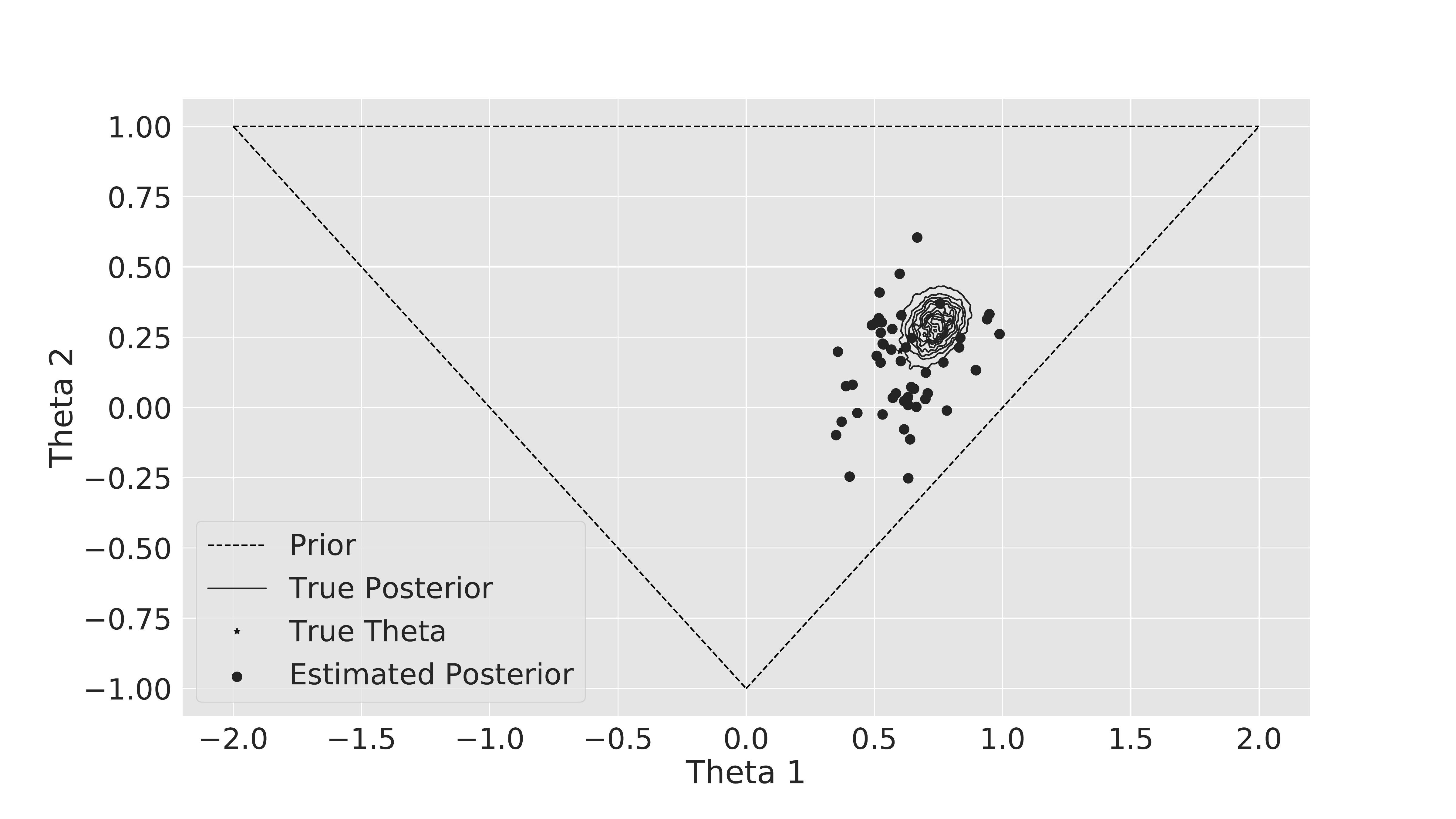}}
\hfill
\subfloat[CNN - training set of $10^4$ samples.]{\label{fig:figure14_6}\includegraphics[width=.3\linewidth]{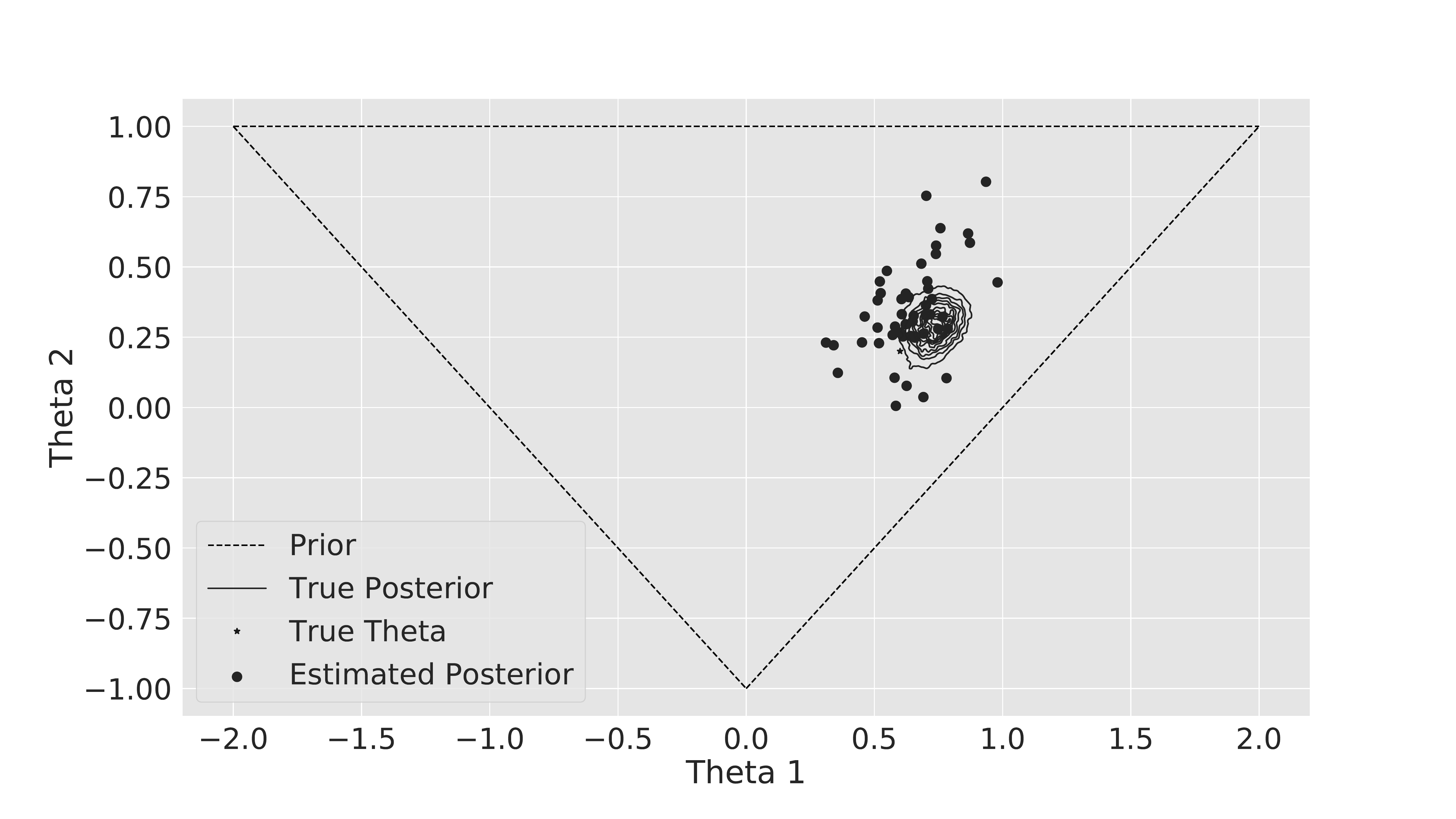}}

\subfloat[DNN - training set of $10^5$ samples.]{\label{fig:figure14_7}\includegraphics[width=.3\linewidth]{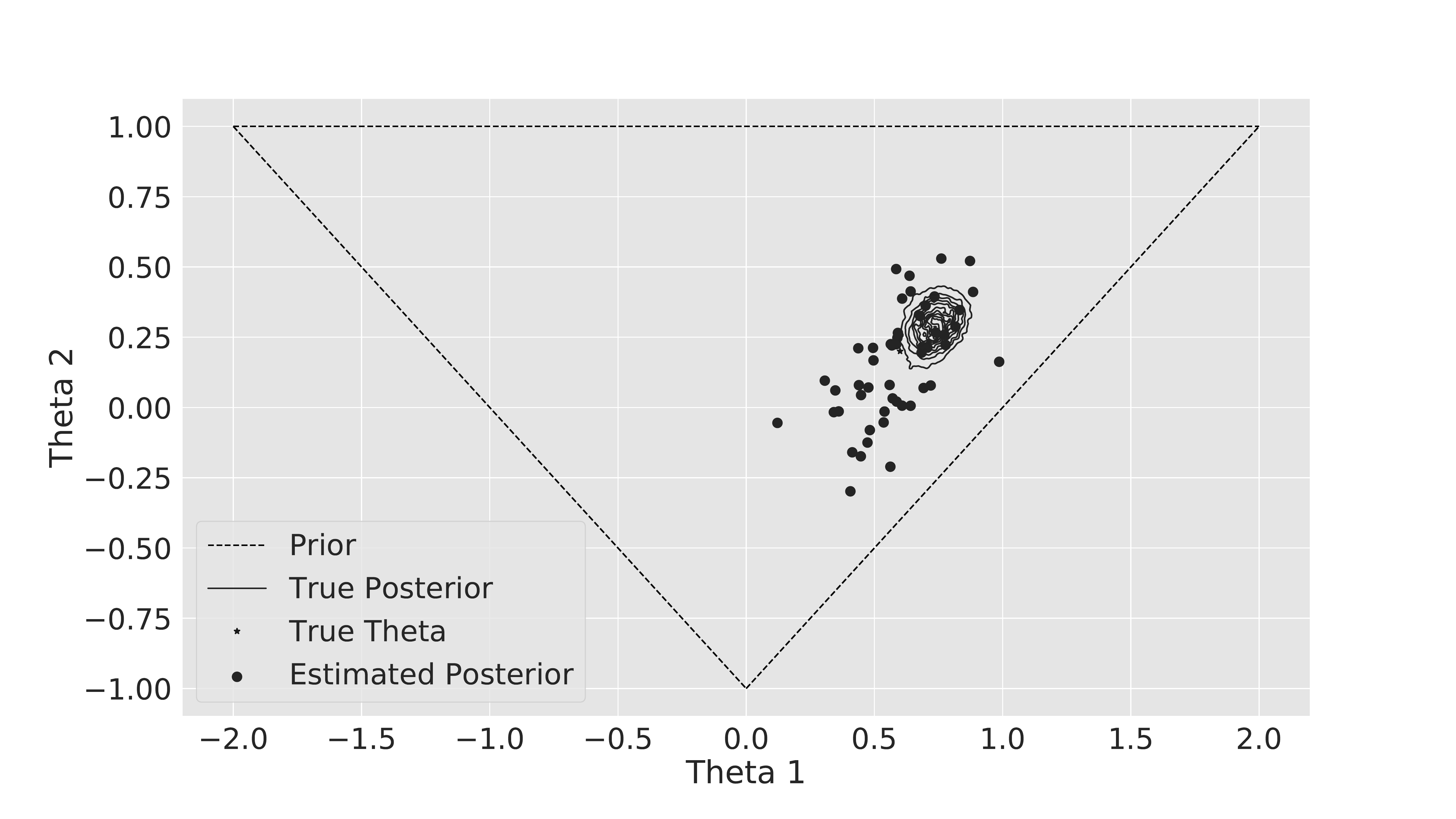}}
\hfill
\subfloat[PEN - training set of $10^5$ samples.]{\label{fig:figure14_8}\includegraphics[width=.3\linewidth]{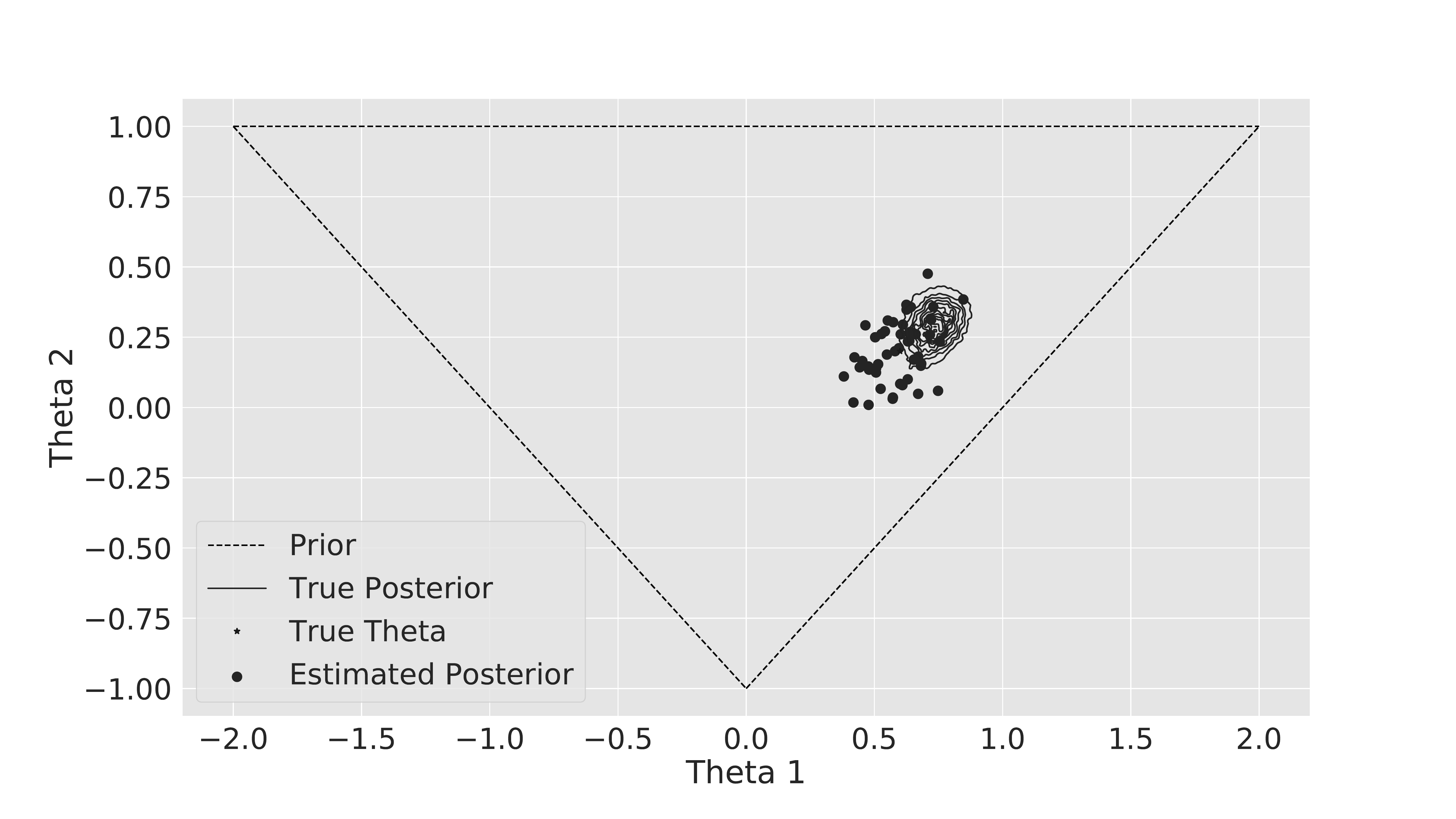}}
\hfill
\subfloat[CNN - training set of $10^5$ samples.]{\label{fig:figure14_9}\includegraphics[width=.3\linewidth]{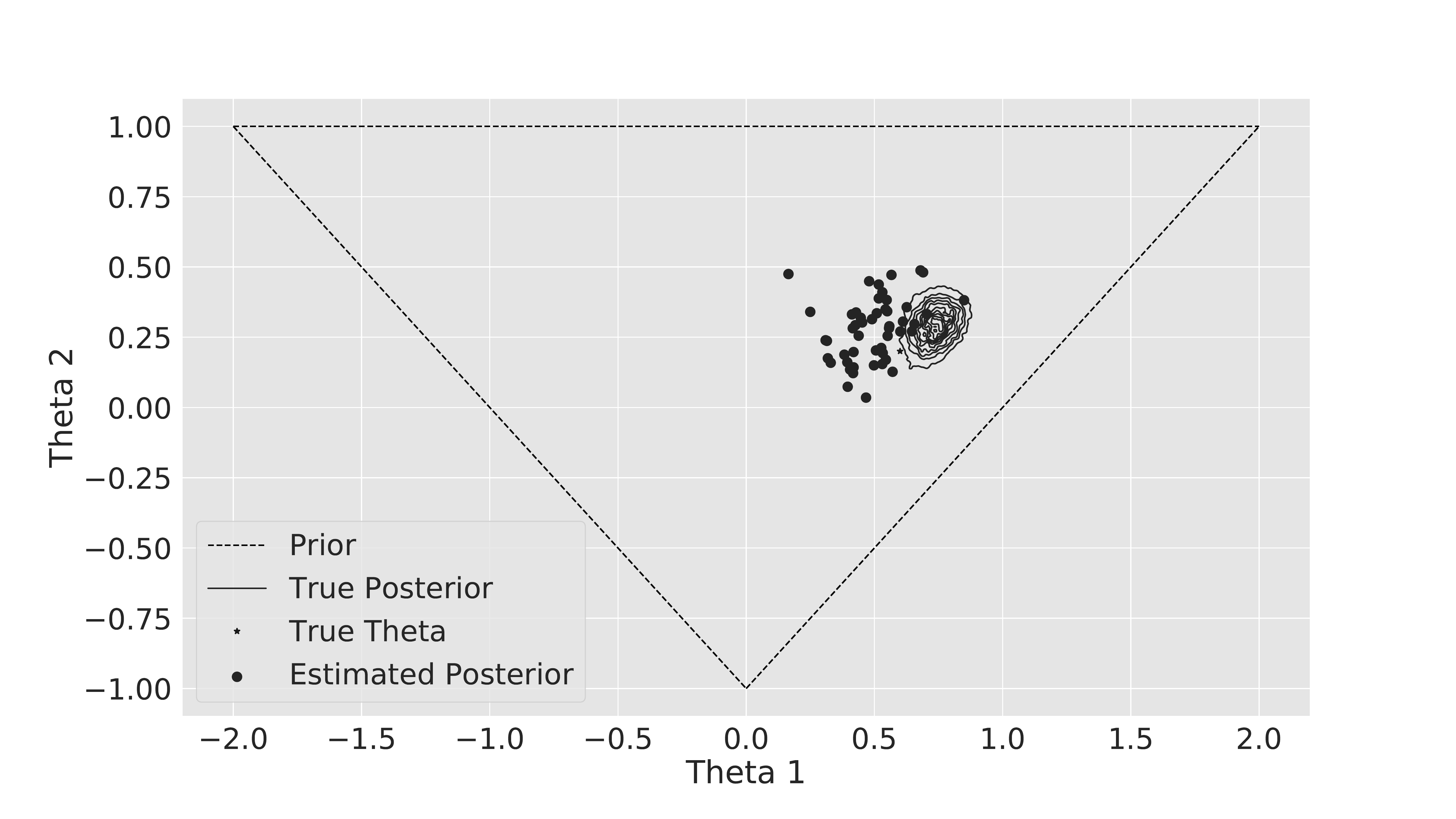}}

\caption{MA2 model: Estimated posterior compared to true posterior for each ANN architecture.}
\label{fig:ma2_posteriors}
\end{figure*}

A visual comparison of estimated posteriors is shown in Fig. \ref{fig:ma2_posteriors}. In order to estimate the posterior, the ABC reference table method was used with $0.01 \%$ acceptance ratio (50 samples accepted out of $5 \times 10^5$ trials). The training, test, validation and ABC trial data samples were consistent and the same across different architectures. In order to calculate the exact posterior distribution, the Random Walk Metropolis-Hastings method was used. Kernel Density Estimation (KDE) was used to visualize the exact posterior.

The posterior estimates in Fig. \ref{fig:ma2_posteriors} reflect comparable performance between PEN and CNN architectures for larger training set sizes. The DNN architecture in comparison is less data-efficient with the posterior estimates showing larger variation from the true posterior. It can also be observed that $10^4$ training samples are enough for the PEN and CNN architectures to be used as accurate high-quality summary statistics.

\subsection{The Lotka-Volterra Model}
\label{sec:lv}
\begin{figure*}
\subfloat[DNN - round 2. Trial count 4591.]{\label{fig:figure14_1}\includegraphics[width=.3\linewidth]{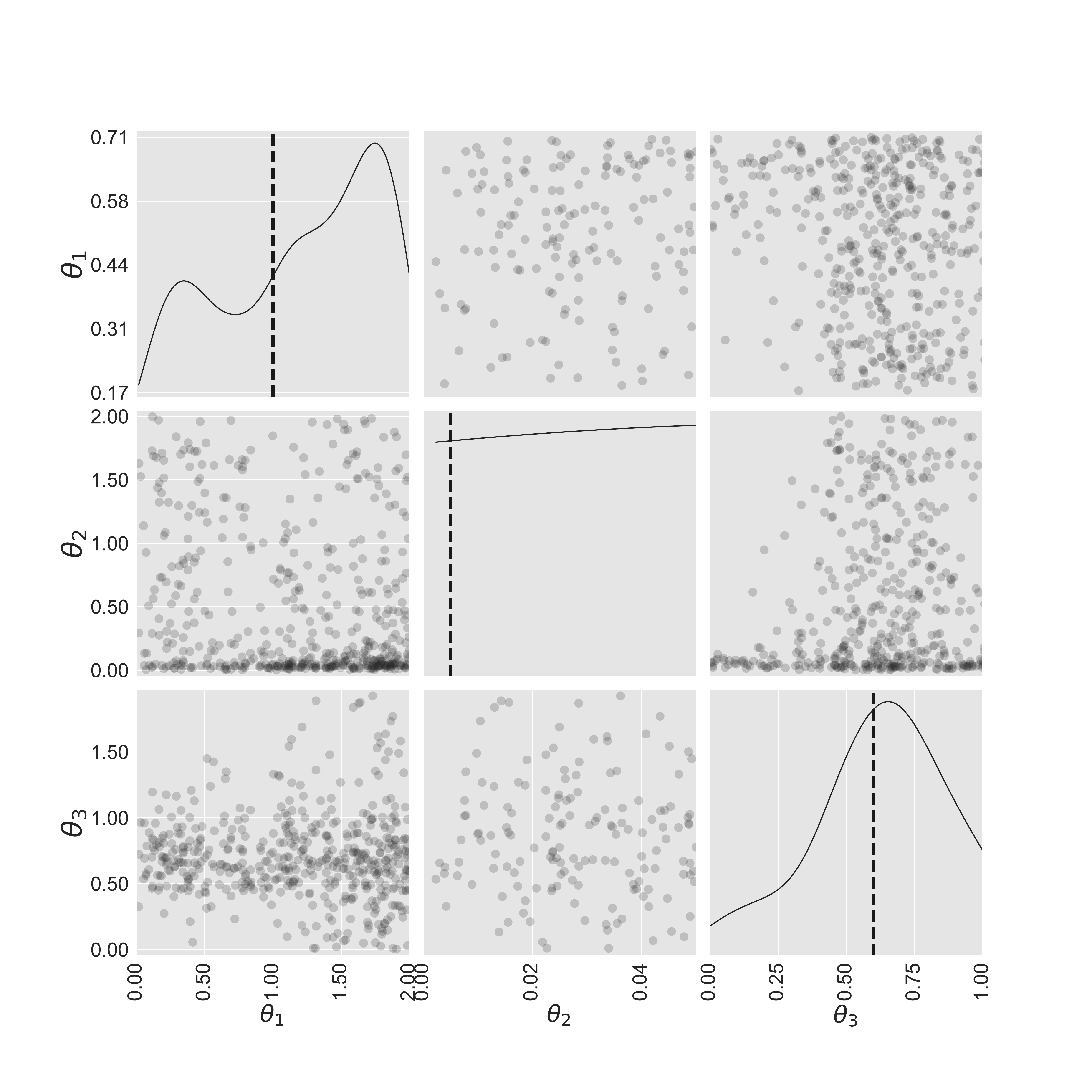}}
\hfill
\subfloat[DNN - round 4. Trial count 23153.]{\label{fig:figure14_2}\includegraphics[width=.3\linewidth]{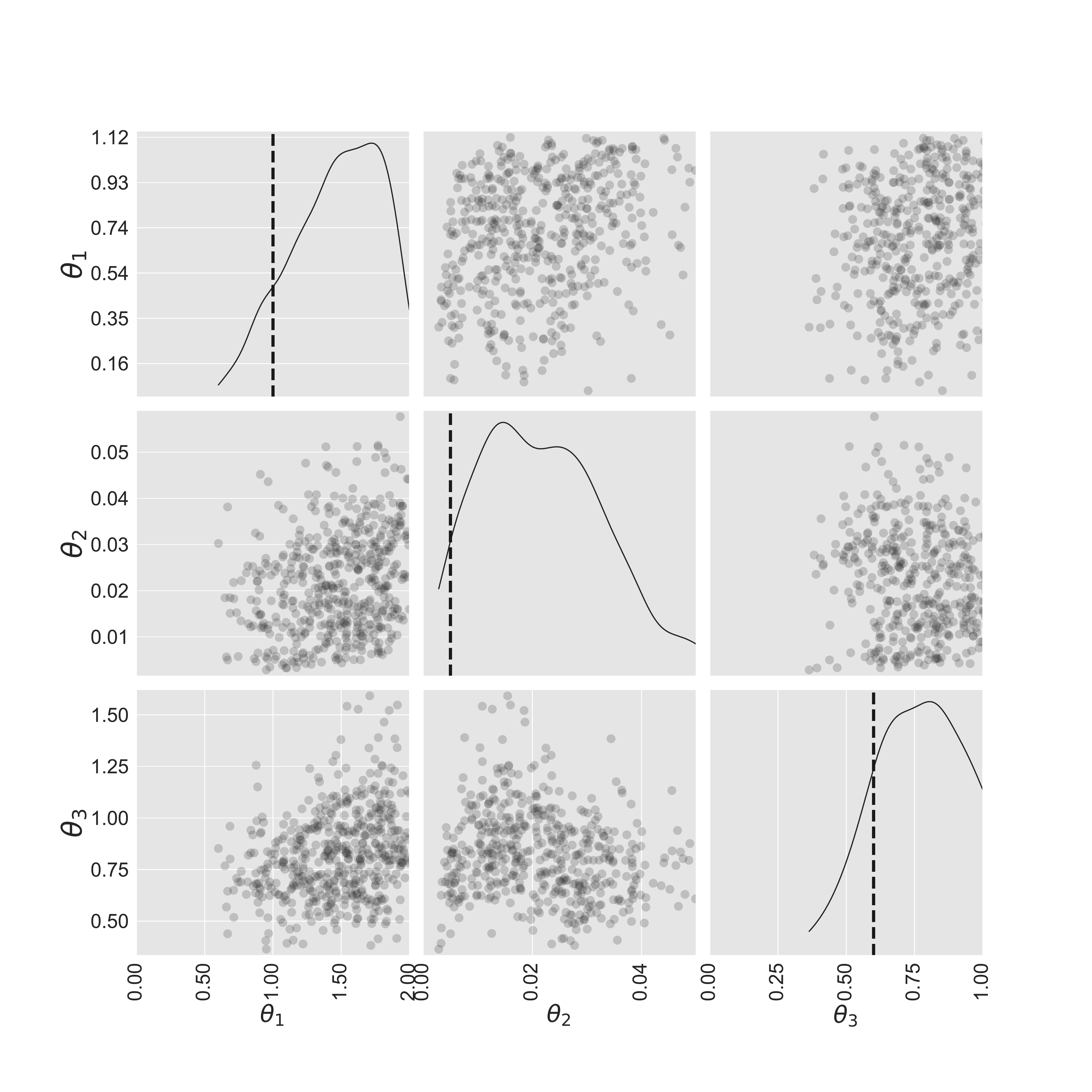}}
\hfill
\subfloat[DNN - round 6. Trial count 302865.]{\label{fig:figure14_3}\includegraphics[width=.3\linewidth]{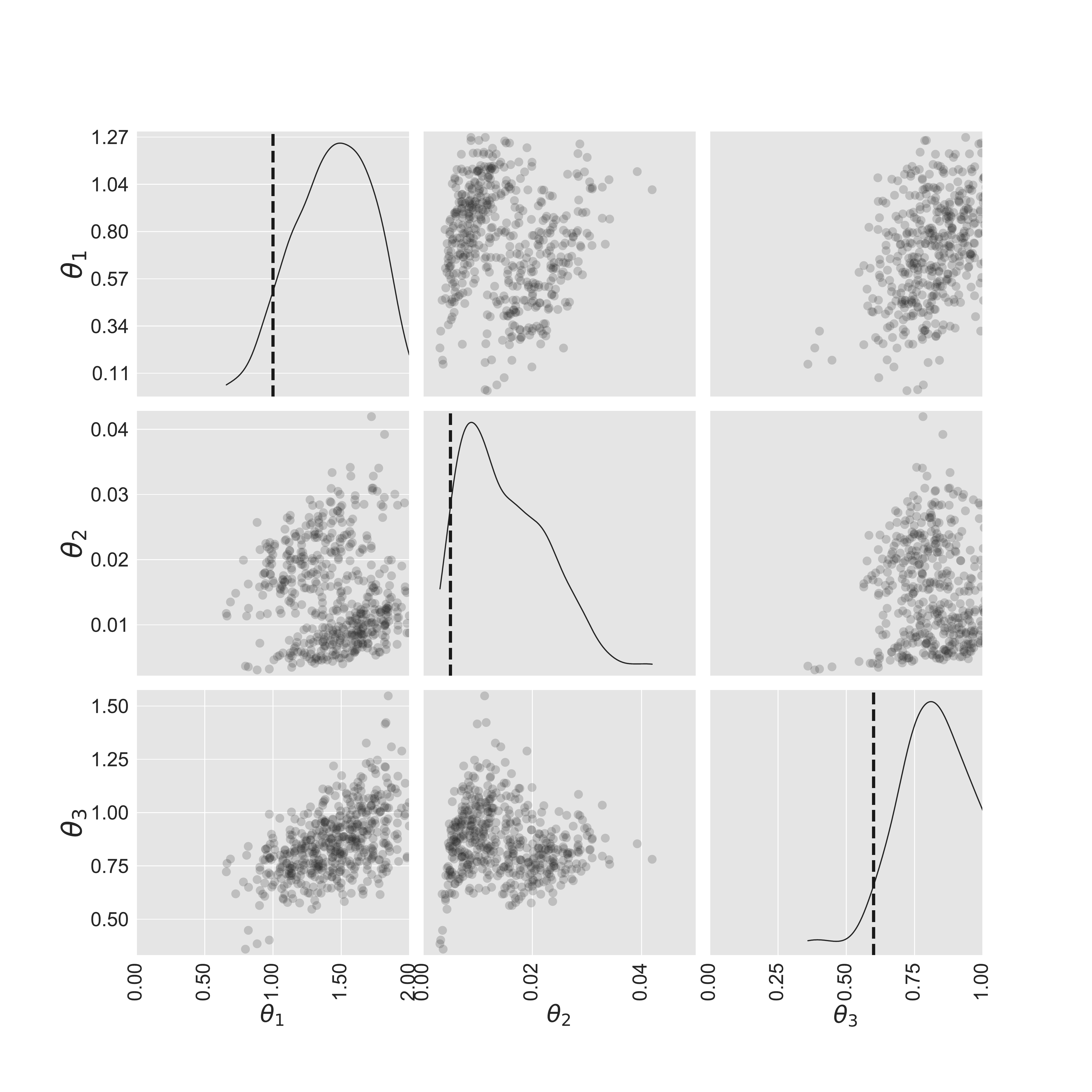}}\\[-2.4ex]
\subfloat[PEN - round 2. Trial count 4645.]{\label{fig:figure14_7}\includegraphics[width=.3\linewidth]{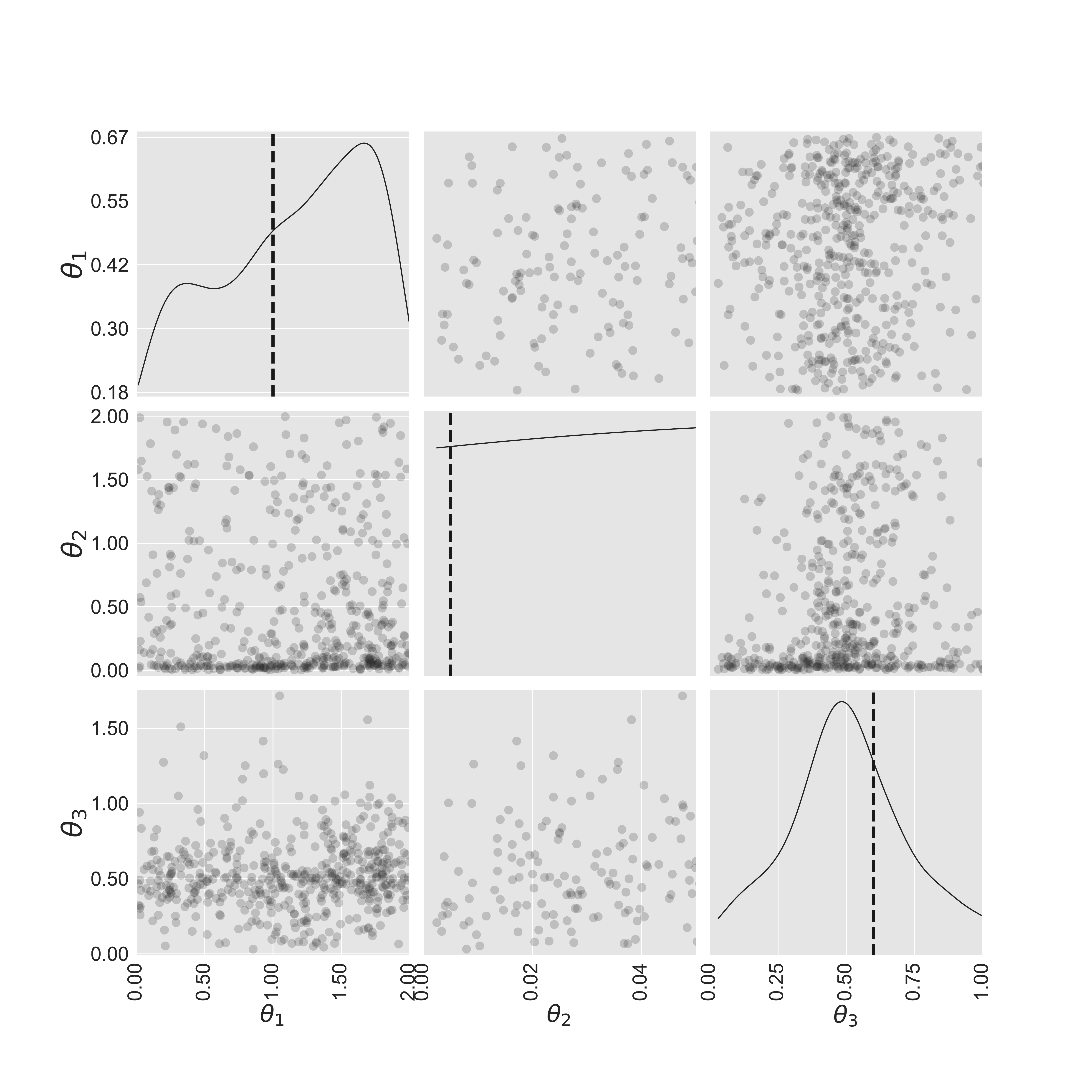}}
\hfill
\subfloat[PEN - round 4. Trial count 25753.]{\label{fig:figure14_8}\includegraphics[width=.3\linewidth]{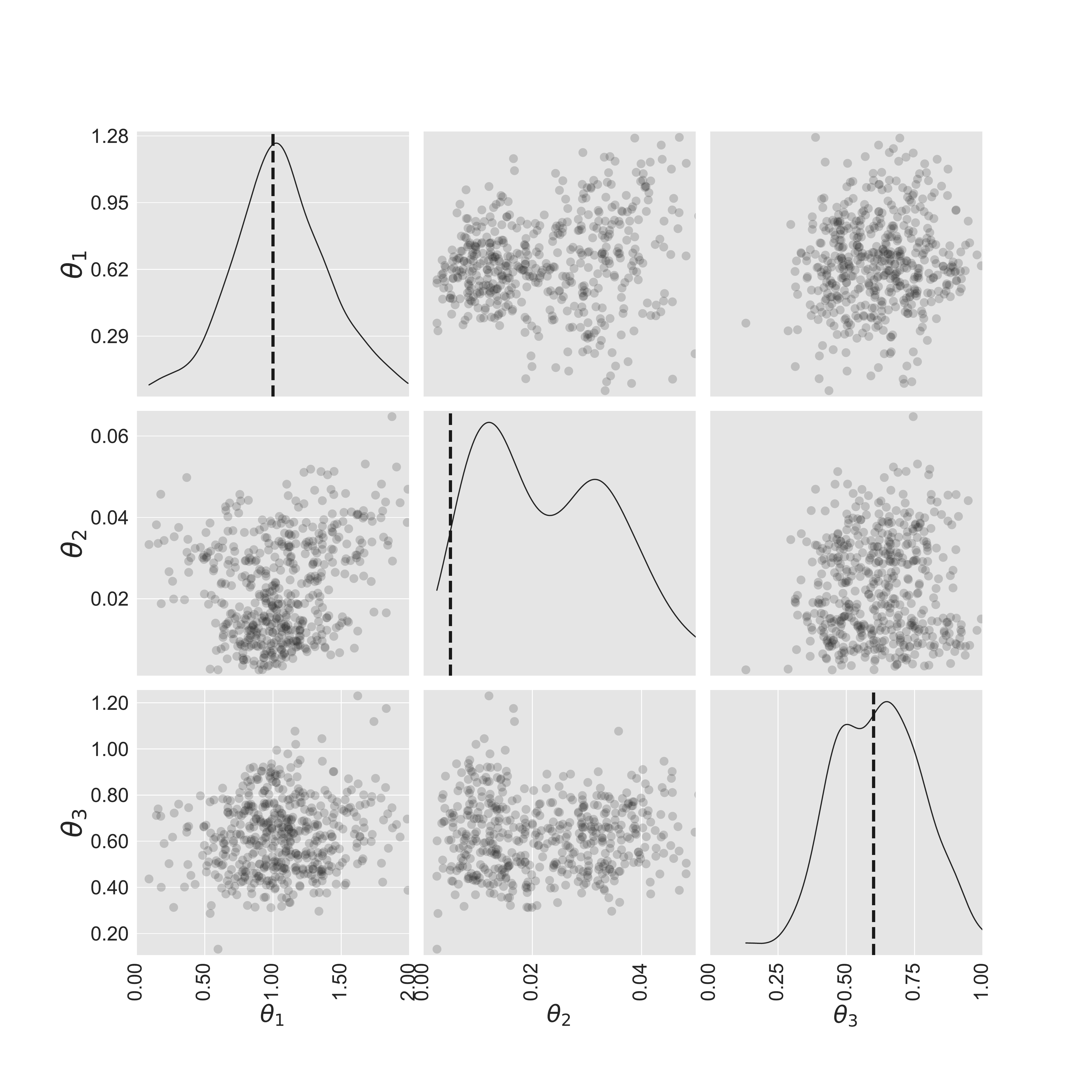}}
\hfill
\subfloat[PEN - round 6. Trial count 261073.]{\label{fig:figure14_9}\includegraphics[width=.3\linewidth]{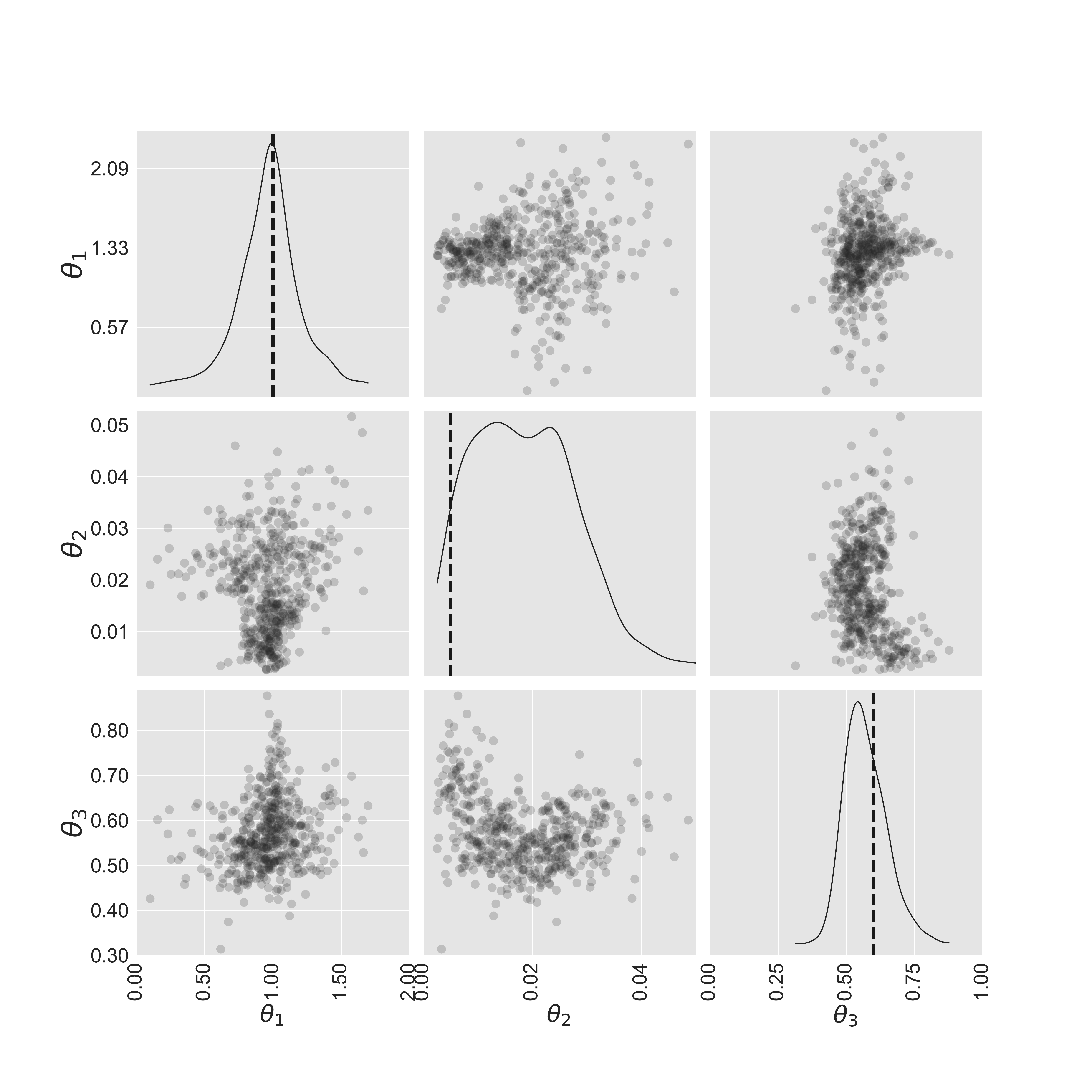}}\\[-2.4ex]
\subfloat[CNN - round 2. Trial count 4207.]{\label{fig:figure14_4}\includegraphics[width=.3\linewidth]{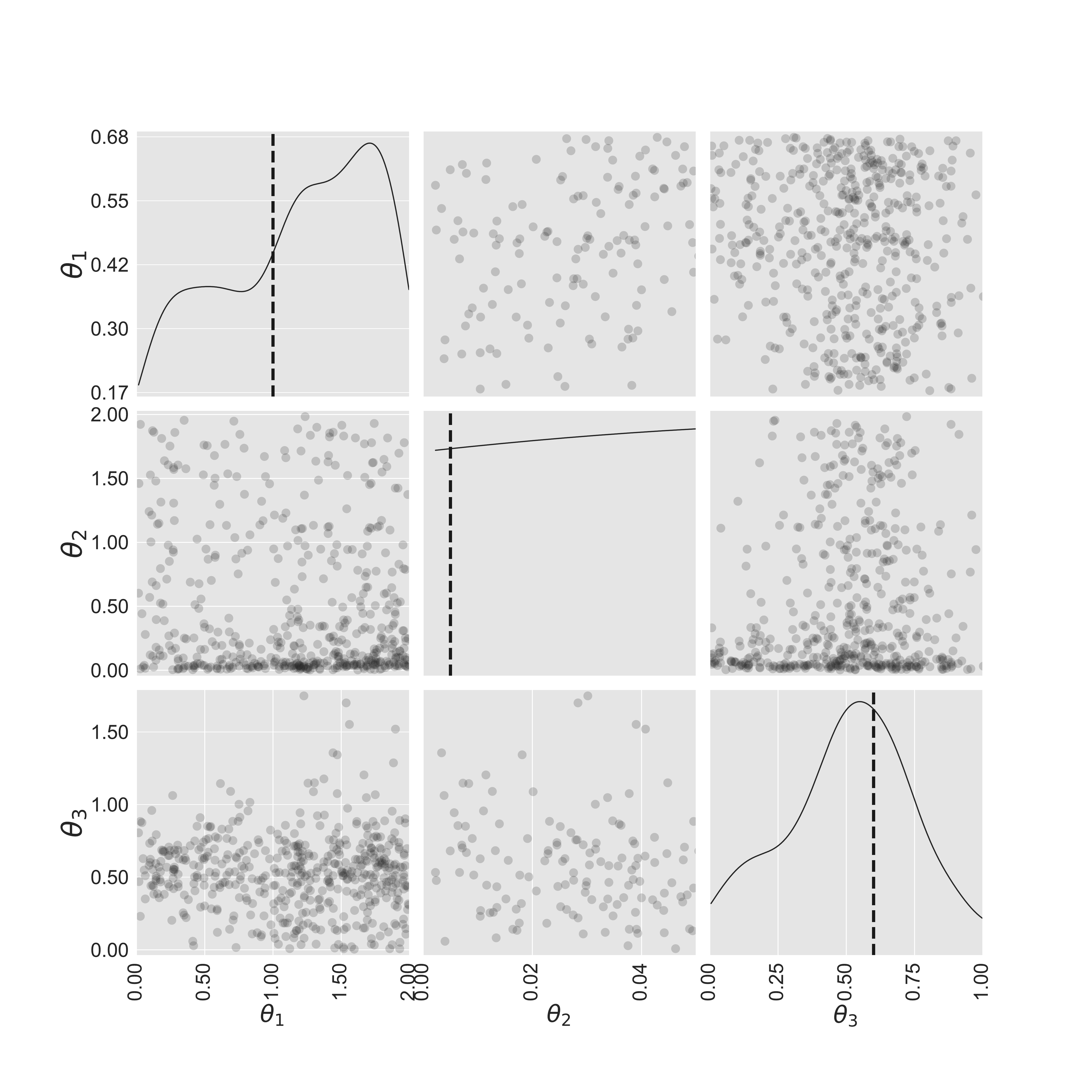}}
\hfill
\subfloat[CNN - round 4. Trial count 20061.]{\label{fig:figure14_5}\includegraphics[width=.3\linewidth]{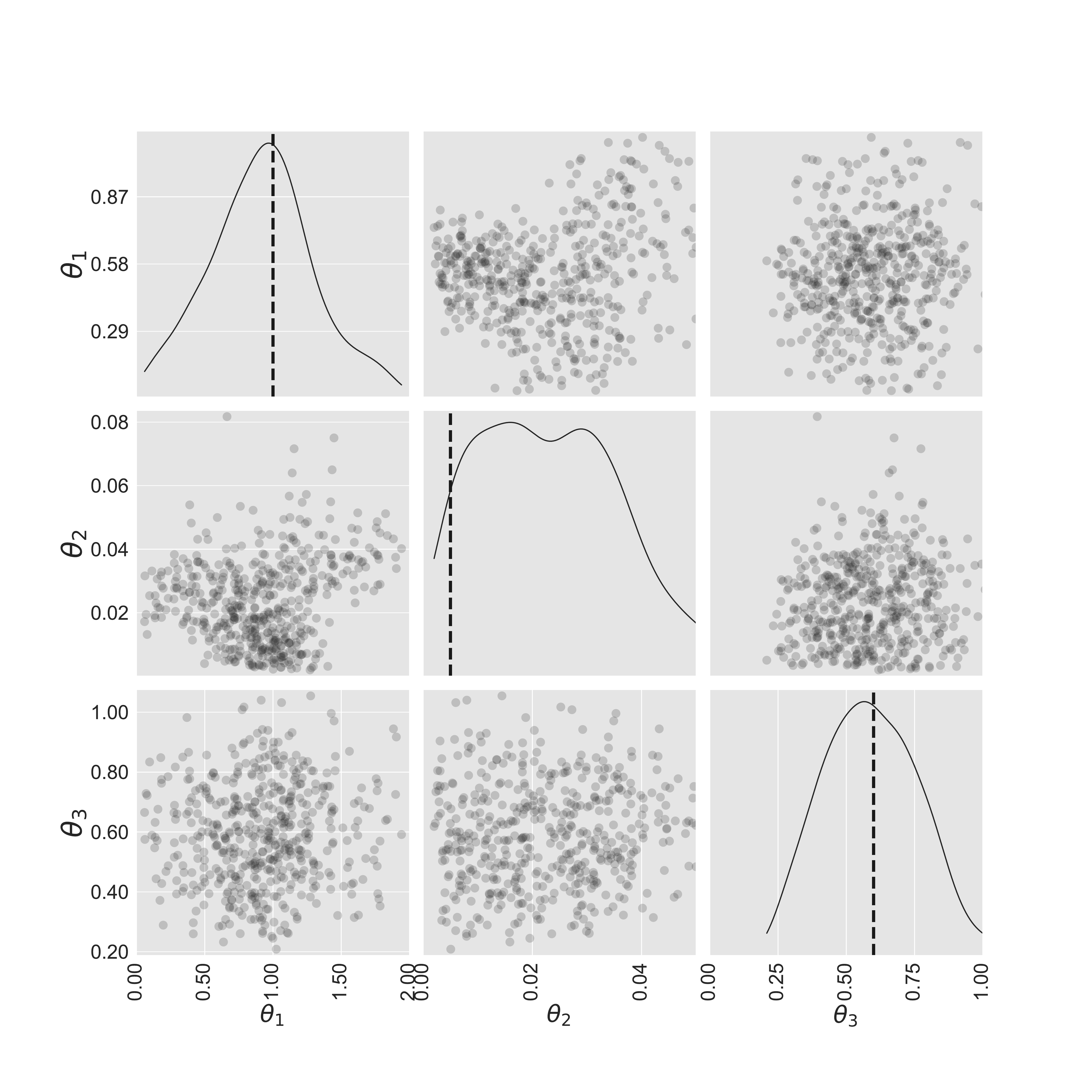}}
\hfill
\subfloat[CNN - round 6. Trial count 221274.]{\label{fig:figure14_6}\includegraphics[width=.3\linewidth]{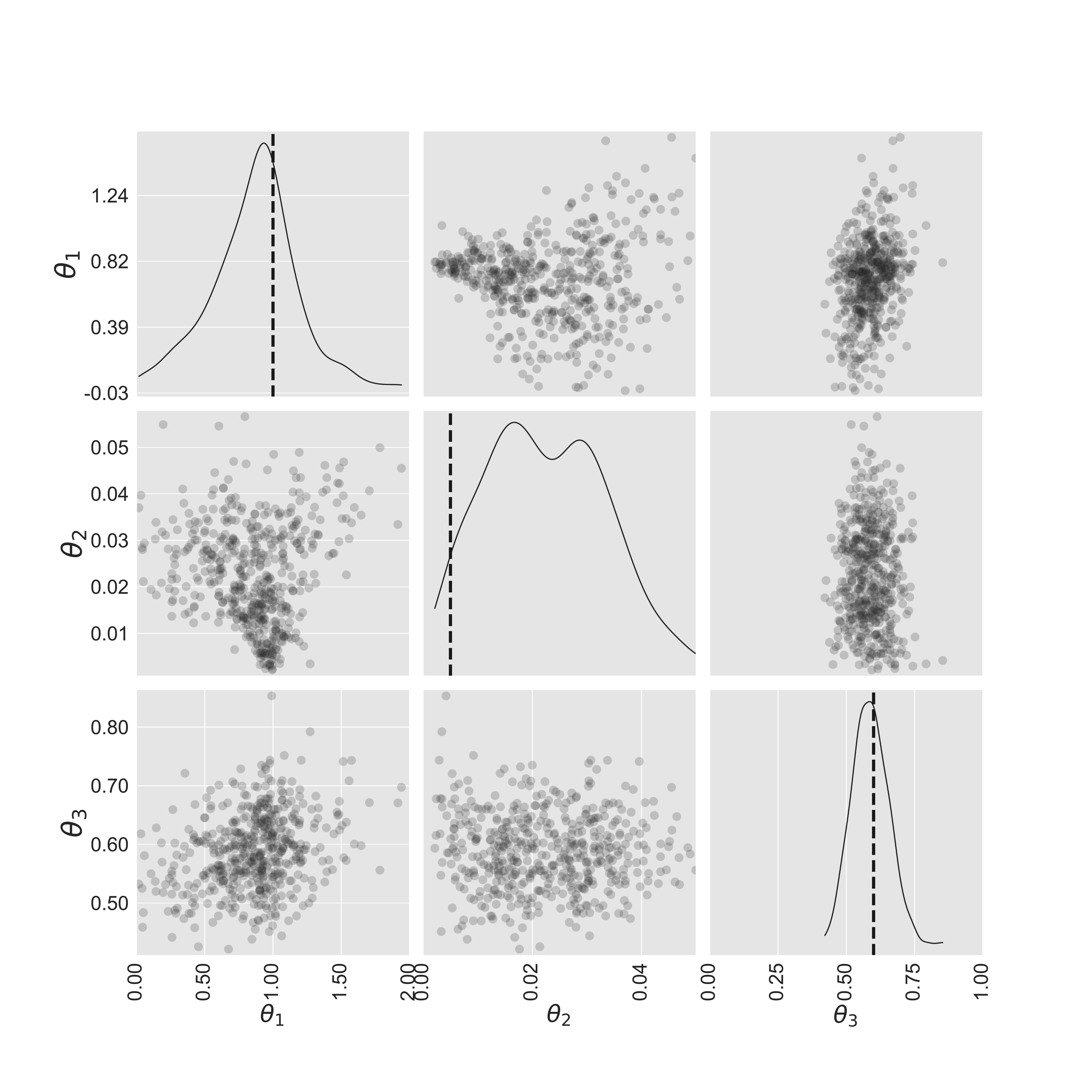}}\\[-2.4ex]
\subfloat[Raw TS - round 2. Trial count 2885.]{\label{fig:figure14_4}\includegraphics[width=.3\linewidth]{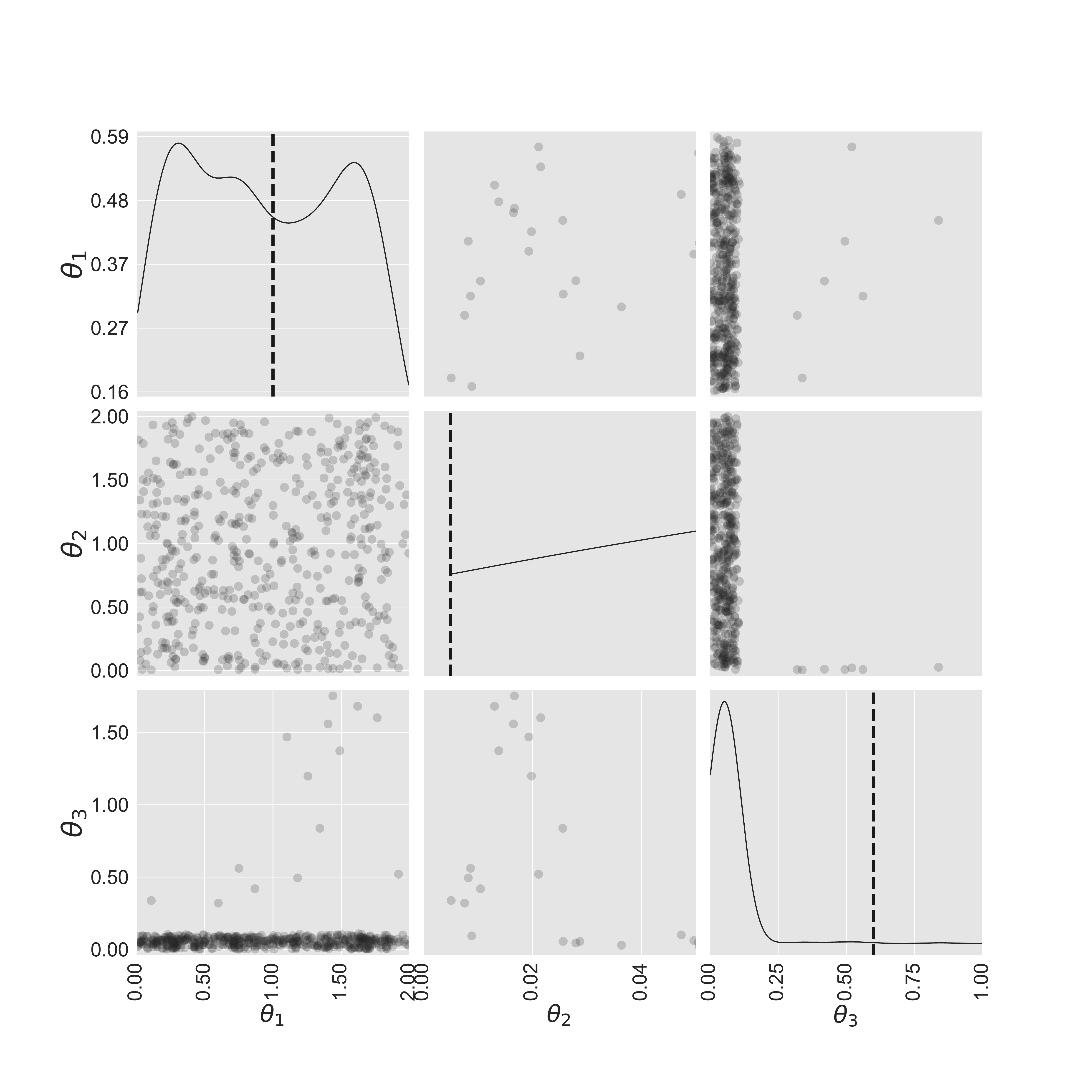}}
\hfill
\subfloat[Raw TS - round 4. Trial count 34250.]{\label{fig:figure14_5}\includegraphics[width=.3\linewidth]{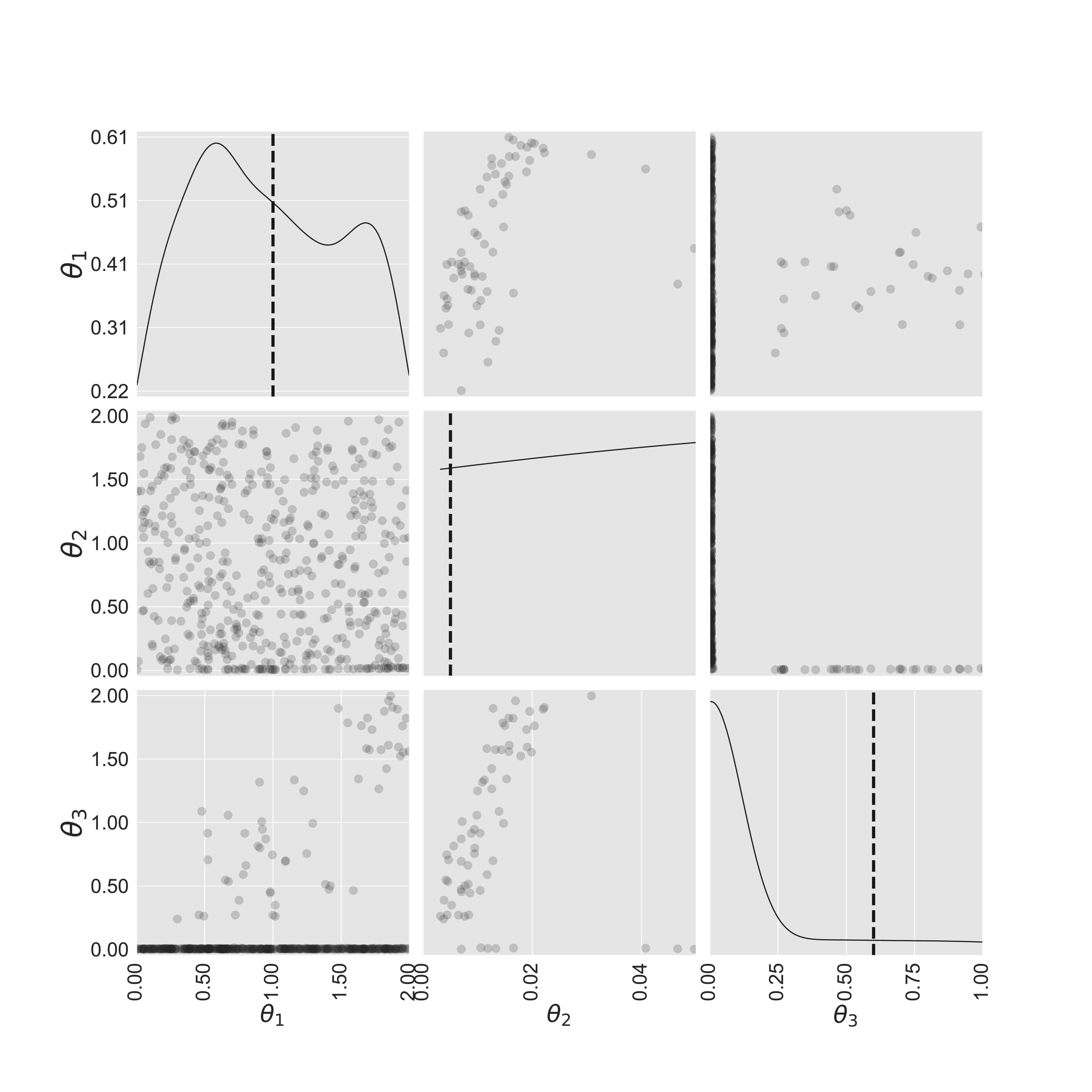}}
\hfill
\subfloat[Raw TS - round 6. Trial count 241228.]{\label{fig:figure14_6}\includegraphics[width=.3\linewidth]{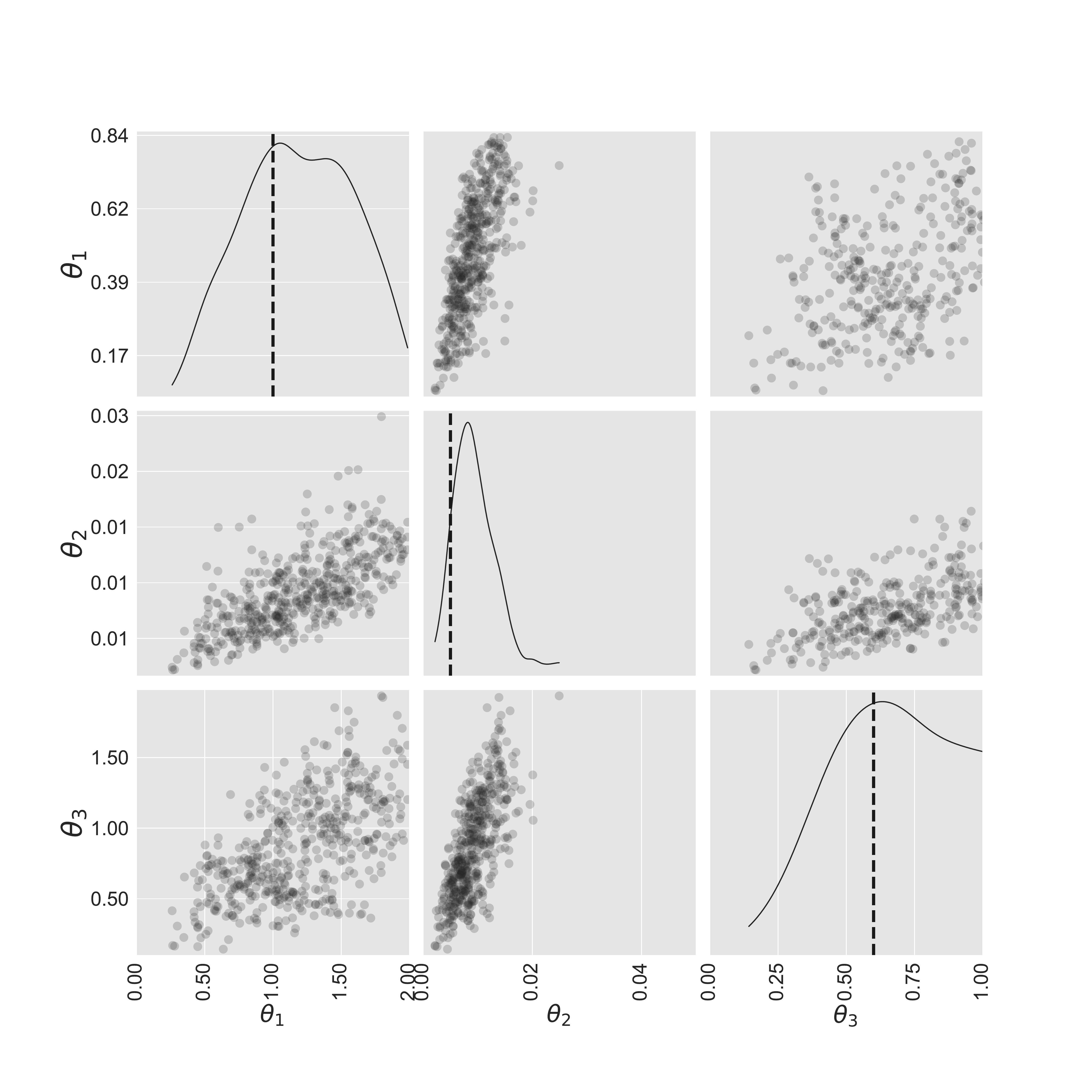}}

\caption{Lotka-Volterra model: Estimated posterior with SMC-ABC using different architectures as summary statistics, and using no summary statistics (raw time series as statistics - denoted as Raw TS above).}
\label{fig:lv_posteriors}
\end{figure*}

The Lotka-Volterra model describes predator-prey population dynamics and is a popular likelihood-free test problem. Here we consider a model variant characterized as a stochastic Markov jump process \cite{prangle2017adapting} simulated using the stochastic simulation algorithm (SSA) \cite{gillespie1977exact}. The model consists of three events - prey reproduction, predation (predator hunts prey and takes part in reproduction) and predator death. The following equations describe the three events,
\begin{align*}
    \mathcal{X}_1 &\longrightarrow 2\mathcal{X}_1,\\
    \mathcal{X}_1 + \mathcal{X}_2 &\longrightarrow 2\mathcal{X}_2,\\
    \mathcal{X}_2 &\longrightarrow \phi.
\end{align*}

The parameters ${\bf \theta} = \{ \theta_1, \theta_2, \theta_3\}$ control the three events described above with rates $\theta_1 \mathcal{X}_1, \theta_2 \mathcal{X}_1 \mathcal{X}_2, \theta_3 \mathcal{X}_2$ respectively. The initial conditions of the model are set to be $\mathcal{X}_1=50$ and $\mathcal{X}_2=100$. Each time series consists of 30 observations from $t=0$ till $t=30$ with a resolution of 1 time step. The true parameters of the inference problem are $[1.0, 0.005, 0.6]$. In certain regions of the parameter space where $\theta_1$ is large and $\theta_2$ is small, the prey population can grow (or \emph{explode}) to a very large value, causing extremely long simulation times. In order to mitigate the effect of such samples, a simulation timeout value (1 second) is specified to the stochastic simulation algorithm (SSA) solver. The samples that exceed the timeout duration are discarded, and new samples from the prior are substituted in exchange.

The training data are sampled uniformly in the interval [0.005, 6.0] for all three parameters $\theta_1, \theta_2$ and $\theta_3$. The training set size is varied in $[3 \times 10^4, 10^5, 2 \times 10^5, 5 \times 10^5]$. The DNN and $\text{PEN}_{10}$ architectures follow description from \cite{10.2307/26384090} and \cite{2019arXiv190110230W} respectively. Both species - predator and prey take part in the parameter inference process. Model training \textit{approach 2} (Sec \ref{sec:model_training}) is used in all experiments for this test problem.

\begin{table}
\small
\centering
\scalebox{1}{%
\setlength{\tabcolsep}{10pt}
\begin{tabular}{c|c|c|c|c}
\toprule
$\text{Network}$ &  $3 \times 10^4$ & $10^5$ & $2 \times 10^5$ & $5 \times 10^5$ \\
\midrule
$CNN$ &   ${\bf 0.727 \pm 0.005}$ &  ${\bf 0.719 \pm 0.002}$ &  ${\bf 0.717 \pm 0.002}$ &  ${\bf 0.717 \pm 0.002}$ \\
$PEN_{10}$ &  $0.823 \pm 0.043$ &  $0.785 \pm 0.036$ &  $0.757 \pm 0.020$ & $0.756 \pm 0.023$ \\
$DNN$ &  $0.857 \pm 0.027$ &  $0.866 \pm 0.017$ & $0.878 \pm 0.019$ & $0.925 \pm 0.041$ \\
\end{tabular}}
\caption{$E_{\%}$ metric calculated on independent random test sets for different architectures on the Lotka-Volterra model for training set sizes $x \times 10^4 - 5 \times 10^5$. The values represent the mean and standard deviation over 10 independent experiments.}
\label{tab:lv_ds}
\end{table}

Table \ref{tab:lv_ds} compares the informativeness (in terms of the $E_{\%}$ measure) of the three ANN architectures over varying training set sizes in [$3 \times 10^4 - 5 \times 10^5$]. The values represent the mean and standard deviation of $E_{\%}$ values over 10 distinct repetitions. Each repetition consists of a uniformly sampled training set (of the size specified in the table), a uniformly sampled validation set of $2 \times 10^4$ samples used in the training process, and a uniformly sampled test set of $10^5$ samples used to calculate the $E_{\%}$ values in Table \ref{tab:lv_ds}. The training, validation and test data used in each repetition is consistent and the same for all 3 ANN architectures to enable a fair comparison.

It is observed that the CNN architecture is consistently more informative as compared to the PEN and DNN architectures. The difference in learning capabilities is starkest when the training set size is most limited, i.e., $3 \times 10^4$ samples. As training set size increases, the gap in learning capabilities of the CNN and PEN architectures narrows. The DNN struggles in comparison due to the nature of the data. The oscillatory nature of time series' and local patterns from the Lotka-Volterra model are better characterized and represented as features by the CNN and PEN architectures. In case of the CNN, the convolutional window enables learning local patterns and discriminating time series features, while the PEN architecture leverages partial (local) exchangeability to accomplish the same. The DNN is fully connected, and not conditioned on subsets of the input time series - therefore, it is unable to learn local informative patterns.

Figure \ref{fig:lv_posteriors} depicts the estimated posterior distributions of $500$ samples each, inferred using Sequential Monte Carlo - ABC (SMC-ABC). ANN architectures trained using $2 \times 10^5$ samples were used as summary statistics for SMC-ABC. In addition, comparison without using any summary statistics, i.e., using raw time series is also included. The tolerance thresholds ($\epsilon$'s) are selected using a relative scheme, where the $20$-th percentile value of the ABC distances from the previous round is selected as $\epsilon$.  It can be observed that the CNN architecture performs slightly better than $\text{PEN}_{10}$ in the inference task for $\theta_3$, while $\text{PEN}_{10}$ performs slightly better in the case of $\theta_1$. Using the raw time series results in the best inference performance in the final SMC-ABC generation for inferring $\theta_2$, with the DNN being next-best. It is also interesting to note that the ANN architectures deliver better inference performance as compared to not using any statistic (raw time series) in generation 4. The $\text{PEN}_{10}$ and CNN architectures are found to perform closely across generations.


\subsection{A high-dimensional genetic oscillator}
\begin{figure*}
\subfloat[Trajectories of genetic oscillator]{\label{fig:vilar_ts}\includegraphics[width=.60\linewidth]{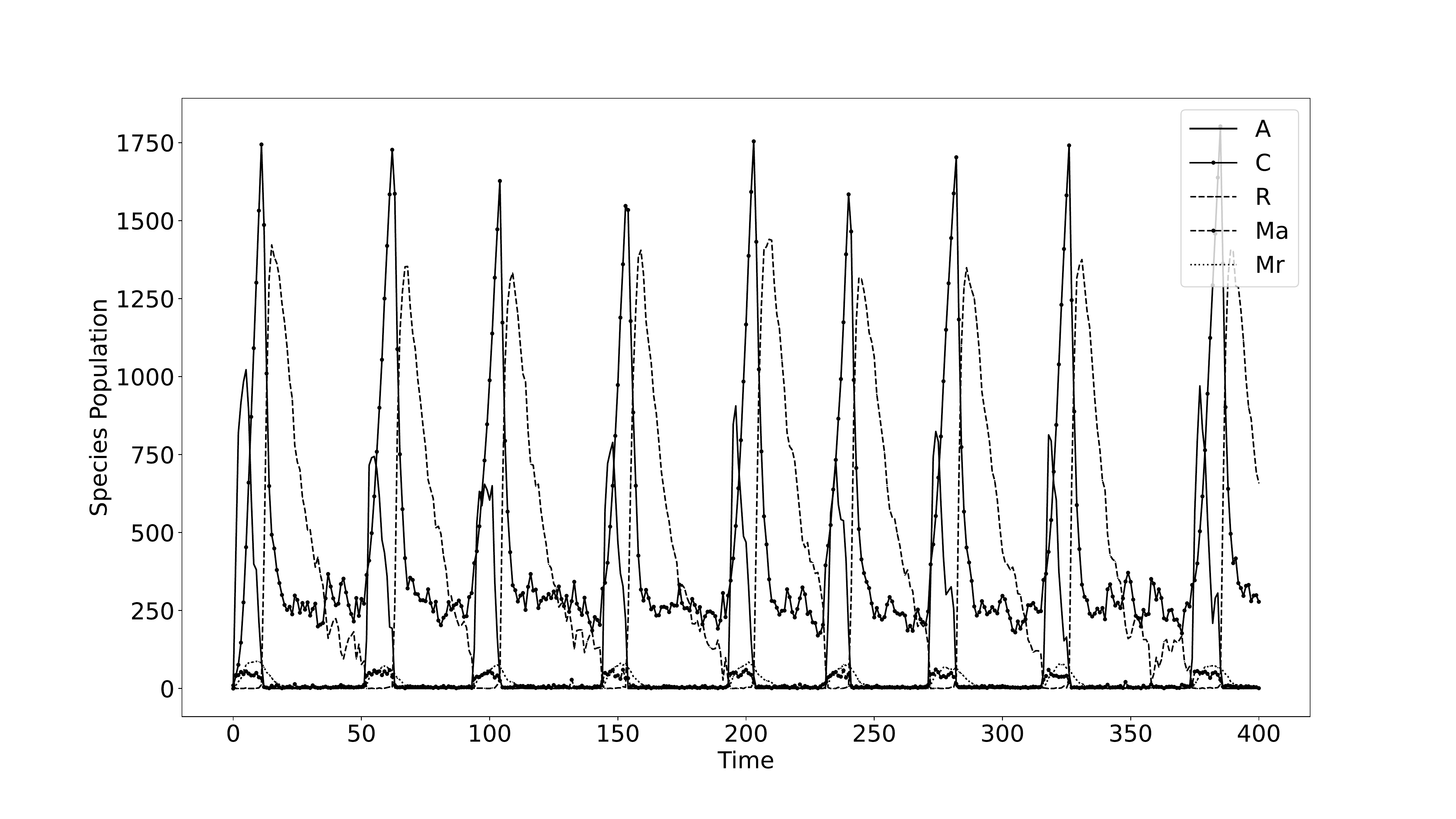}}
\hfill
\subfloat[Biochemical network]{\label{fig:vilar_net}\includegraphics[width=.39\linewidth]{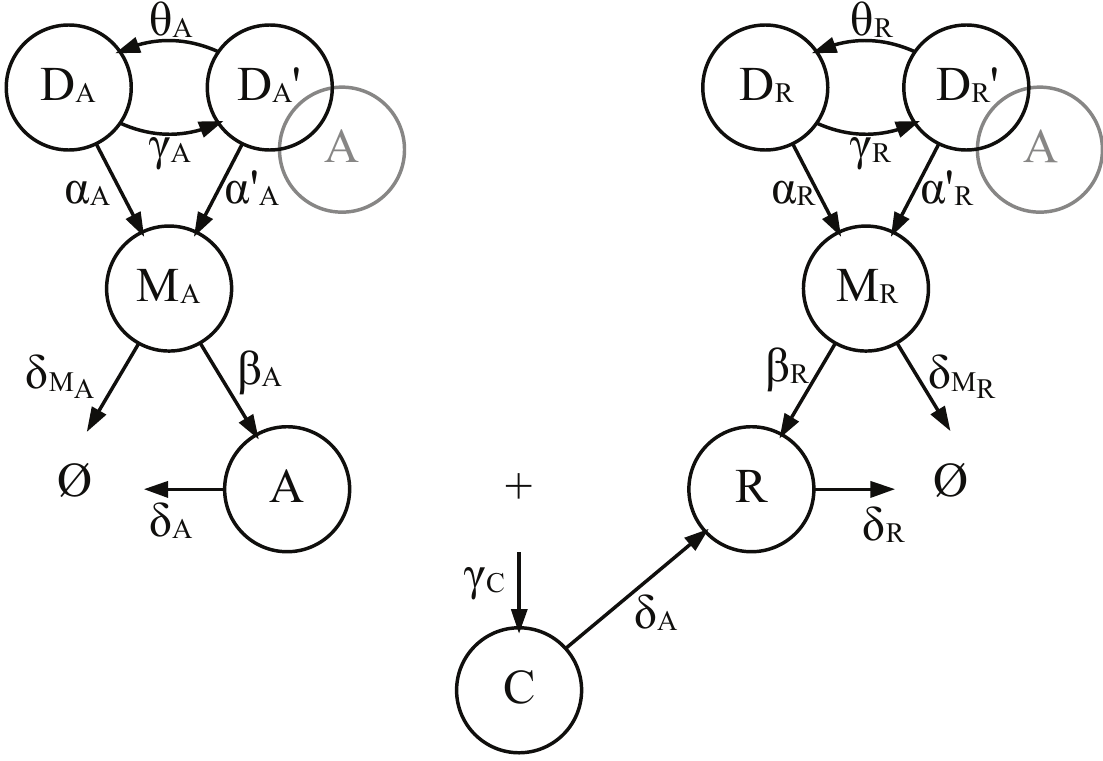}}
\caption{(a) Time series responses corresponding to all mRNA and proteins. The model is simulated at the well-known reference point \cite{vilar2002mechanisms} with the time vector being $t = \{0: 200: 0.5\}$. The plot was generated using the stochastic simulation service (StochSS) package \cite{drawert2016stochastic}. (b) The network structure of the genetic oscillator, see text for details.}
\label{fig:vilar_info}
\end{figure*}

We next consider a complex, high-dimensional biochemical reaction network with oscillatory behavior \cite{vilar2002mechanisms}. The network involves $9$ species undergoing $18$ reactions parameterized by $15$ reaction constants (Figure \ref{fig:vilar_net}). The model is a gene regulatory network based on a positive-negative feedback loop mimicking a circadian clock where the activator protein $A$ binds to the corresponding gene promotor site to up-regulate transcription, but it also activates transcription of a repressor protein $R$ which in turn reacts with $A$ to form a new complex $C$, thus sequestering the activator. This model was one of the first realistic gene regulatory models to highlight the impact of intrinsic noise due to low copy numbers of the species. In particular, the system's dynamics are robust under intrinsic noise in the chemical reactions, and in fact, the model suggests an increased robustness to fluctuations in parameters as compared to deterministic models using ordinary differential equations. To incorporate intrinsic noise, the model is realized as a continuous-time discrete space Markov chain where the probability of a reaction occurring at a certain state of the system is governed by the chemical master equation. 

Fig \ref{fig:vilar_info}(b) depicts the biochemical reaction network. $D_A$ and $D_A'$ correspond to the copy numbers of the activator gene with and without A bound to its promoter, respectively. The same applies for $D_R$ and $D_R'$ for the repressor gene. Transcription rates to mRNA ($M_R$ and $M_R$) are denoted by $\alpha$ parameters, while translation rate parameters into activator and repressor proteins are  denoted by $\beta$. Other parameters, $\delta$ denote rates of spontaneous degradation, $\gamma$ the rates of binding of $A$ to other species, and $\theta$ denotes the rates of unbinding of A from those species. Finally, a complex $C$ is formed by the reaction between $A$ and $R$. The set of chemical reactions are

\allowdisplaybreaks
\begin{equation*}
\label{eq:rxns}
\begin{aligned}
D_A^* &\xrightarrow{\theta_A} D_A,\\
D_A,A &\xrightarrow{\gamma_A} D_A^*,\\
D_R^* &\xrightarrow{\theta_R} D_R,\\
D_R,A &\xrightarrow{\gamma_R}D_R^*,\\
D_A^* &\xrightarrow{\alpha_A^*} D_A^*,M_A,\\
D_A &\xrightarrow{\alpha_A} D_A, M_A,\\
M_A &\xrightarrow{\delta_{MA}} \phi,\\
M_A &\xrightarrow{\beta_A} A, M_A,\\
D_A^* &\xrightarrow{{\theta}_A} D_A^*,A,\\
\end{aligned} \quad
\begin{aligned}
D_R^* &\xrightarrow{{\theta}_A} D_R^*,A,\\
A &\xrightarrow{\delta_A} \phi,\\
A,R &\xrightarrow{\gamma_C} C,\\
D_R^* &\xrightarrow{\alpha_{R^*}} D_R^*, M_R,\\
D_R &\xrightarrow{\alpha_R} D_R, M_R,\\
M_R &\xrightarrow{\delta_{MR}} \phi,\\
M_R &\xrightarrow{\beta_R} M_R,R,\\
R &\xrightarrow{\delta_R} \phi,\\
C &\xrightarrow{\delta_A} R.
\end{aligned}
\end{equation*}

where the range of reaction-rate parameters considered here are found in Table \ref{tab:prior}.Python code implementing the network is part of the StochSS example library \cite{10.1093/bioinformatics/btab061}. For all numerical experiments, we generate synthetic data using GillesPy2 as part of the StochSS suite of tools. 

It is shown in \cite{vilar2002mechanisms} that the change of certain reaction rate parameters have negligible effect on the underlying behavior of the model (oscillations), including degradation rates of mRNAs ($\delta_{M_R}$ and $\delta_{M_A}$) and the translation rates of the main proteins ($\beta_A$ and $\beta_R$). Thus, from an inference point of view these should be harder to infer compared to, for example, the degradation rates of the proteins ($\delta_{R}$ and $\delta_{A}$), which are known to have a large impact on the periodicity of the oscillations. In general we expect reaction rates associated with the low-level dynamics of the system (transcription) to be more difficult to infer if the observable target lack any of the mRNA species. 

The first part of the experiments focuses on the accuracy of the summary statistics/predicted parameters $\hat{\theta}$ over the prior domain. As a baseline, we consider the time series of the single species $\{C\}$ over a uniform prior bounded by $\bold{dmin}$, $\bold{dmax}$ defined in Table \ref{tab:prior}.
\begin{table}
\scriptsize
\centering
\scalebox{1}{%
\setlength{\tabcolsep}{5pt}
\begin{tabular}{c|c|c|c|c|c|c|c|c|c|c|c|c|c|c}
$\alpha_A$ & $\alpha_A'$ & $\alpha_R$ & $\alpha_R'$ & $\beta_A$ & $\beta_R$ & $\delta_{MA}$ & $\delta_{MR}$ & $\delta_A$ & $\delta_R$ & $\gamma_A'$ & $\gamma_R$ & $\gamma_C$ & $\theta_A$ & $\theta_R$ \\
\hline
0 & 100 & 0 & 20 & 10 & 1 & 1 & 0 & 0 & 0 & 0.5 & 0 & 0 & 0 & 0 \\ 
80 & 600 & 4 & 60 & 60 & 7 & 12 & 2 & 3 & 0.7 & 2.5 & 4 & 3 & 70 & 300 \\
\end{tabular}}
\caption{The lower (first row) and upper bounds (second row) of the uniform prior used for the genetic oscillator test problem.}
\label{tab:prior}
\end{table}

The training data consists of $N=3 \times 10^5$ samples, with a validation set of $2 \times 10^4$ samples and a test set of $10^5$ samples. The $E_{\%}$ values in the tables represent the mean over all samples in the test set. Please note that we will also use $E_{\%}$ to measure parameter inference quality in addition to model informativeness. In case of inference quality, $E_{\%}$ measures the information gain from the posterior over the prior. Two ANN configurations are explored in this work: $\text{\bf setup 1}$ (convolutional layers [25, 50, 100], dense layers [100, 100]) trained using \textit{approach 1} (Section \ref{sec:model_training}), and $\text{\bf setup 2}$ (convolutional layers [32, 48, 64, 96], dense layers [400, 400, 400]) trained using \textit{approach 2} (Section \ref{sec:model_training}). Setup 2 involves larger ANN models for all three architectures trained using a computationally more expensive approach in order to evaluate the potential gains in model accuracy. The time vector for simulating the oscillator model is $t = \{0:200:0.5\}$ unless otherwise stated.

To investigate the performance of the approach we conducted a series of numerical experiments.  First, we compare the three architectures in terms of inference accuracy and training cost. Then, for the CNN, we consider a number of scenarios related both to experimental setup and to the cost of simulation to evaluate the potential of the ANN inference approach in practice for a realistic system. Specifically, we vary the observed species and the amount of observed data, and we also look at the impact of the amount of simulated training data on the performance. 
\begin{table}
\scriptsize
\centering
\scalebox{1}{%
\setlength{\tabcolsep}{5pt}
\begin{tabular}{c|c|c|c|c|c|c|c}
Statistic & sum val.  & median & mean & std. dev. & var. & max & burstiness \\
\hline
Frequency & 1 & 6 & 1 & 8 & {\bf 13} & 7 & {\bf 16}\\
\end{tabular}}
\caption{The frequency of selection of each summary statistic over 50 invocations of the AS algorithm.}
\label{tab:as}
\end{table}

\subsubsection{Comparison of the three network architectures}
Table \ref{tab:CompArch} compares the ANN estimated posterior (on a test set of $10^5$ samples) against the established approximate sufficiency (AS) method \cite{joyce2008approximately} in terms of $E_{\%}$. Larger architectures described in setup 2 were used for all 3 ANN architectures with consistent layer size and scale for each layer type. For reference, ABC inference using the complete pool of available summary statistics is also shown. The candidate pool of summary statistics is shown in Table \ref{tab:as} and includes mean, median, sum of values, standard deviation, variance, max and burstiness \cite{goh2008burstiness}. The most frequently selected statistics are variance and burstiness, and were used for performing ABC inference in conjunction with AS for results depicted in Table \ref{tab:CompArch}. It can be seen that no substantial improvement is obtained using AS over using all available traditional summary statistics. The CNN and $\text{PEN}_{10}$ summary statistics however, result in a very significant improvements with the CNN performing the best overall. The results also highlight the advantage of the proposed method (and of estimated posterior mean in general as a summary statistic) in cases where the candidate pool of statistics might not contain sufficient discriminating ability to allow high quality inference. In such cases, using a highly expressive approximator of the posterior mean (such as the CNN) allows automatic learning of high fidelity summary statistics.

As mentioned earlier when introducing the genetic oscillator, we expect kinetic rate parameters associated with the transcriptions to be more difficult to infer when using only e.g species $\{C\}$ as input to the CNN, which is justified in Table \ref{tab:CompArch} (observe the high $E_\%$ for some $\alpha$, $\theta$ and $\gamma$ parameters). However, $\alpha_A$ performs very well. 

Table \ref{tab:CompArch2} lists the training times and model sizes of the different ANN architectures trained using setup 1. The DNN is the fastest but also the least informative of the three architectures. The CNN had also the slowest training time and largest number of trainable parameters. 

Table \ref{tab:ts_step} depicts a test for the three architectures in inferring the parameters based on differing species, time series range and resolution. We wanted to observe the trade-off between inference quality and the resolution of time points and the time range. Setup 1 was used for all 3 ANN architectures. The values represent the mean posterior estimation error ($E_{\%}$) averaged over all $15$ parameters. The proposed CNN architecture delivers inference with the smallest error in an overwhelming majority of cases. 
Overall we also observe that the increase of more molecular species used as input to the ANNs also increases the quality of inference. 

In order to better understand the nature of the two well-performing architectures - CNN and PEN$_{10}$, table \ref{tab:ts_percent_change_endstep} presents the percentage change in $E_{\%}$ between subsequent time series end points. The CNN benefits the most from higher sampling resolution and longer time series length to extract descriptive features. In cases where the observed time series is short ($\leq 50$h) and sparse (time step $\geq 1$), the PEN$_{10}$ architecture is a better choice. In these 2 cases the PEN is able to be more data efficient and exploit partial exchangeability by viewing the time series as sets instead of ordered data. In short time series with large intervals between observations (step sizes), there is not enough detail as ordered data for approaches like the CNN to work effectively. On the other hand, by viewing the sparse and short time series as sets, the PEN$_{10}$ is able to extract partial sets such as oscillating patterns.

\begin{table}
\small
\centering
\scalebox{1}{%
\setlength{\tabcolsep}{10pt}
\begin{tabular}{p{0.8cm}|c|c|c|c|c}
 & \multicolumn{3}{c|}{\shortstack{ Neural network architectures}} & \multicolumn{2}{c}{\shortstack{ Traditional statistics}} \\
\cline{2-6}
Param. & CNN & {PEN$_{10}$} & DNN & All & AS \\
\hline 
\begin{filecontents*}{accuracy_data_different_models_prior6.csv}
$0$,$1$,$2$,$3$,$4$,$5$
$\alpha_{A}$,${\bf 0.392}$, $0.402$, $0.639$, $1.009$, $1.378$
$\alpha_{A}'$,${\bf 0.512}$, $0.532$, $0.744$, $0.975$ , $0.904$
$\alpha_{R}$,${\bf 0.920}$, $0.938$, $0.990$, $1.785$, $1.936$
$\alpha_{R}'$,${\bf 0.758}$, $0.790$, $0.890$, $0.997$, $0.931$
$\beta_{A}$,${\bf 0.505}$, $0.523$, $0.836$, $1.352$, $1.289$
$\beta_{R}$,${\bf 0.490}$, $0.514$ , $0.691$, $1.161$, $1.214$
$\delta_{MA}$,${\bf 0.494}$, $0.507$, $0.769$, $1.815$, $1.652$
$\delta_{MR}$,${\bf 0.403}$, $0.425$, $0.566$, $0.783$, $0.964$
$\delta_{A}$,${\bf 0.255}$, $0.256$, $0.594$, $0.869$, $0.885$
$\delta_{R}$,${\bf 0.366}$, $0.399$ , $0.774$, $0.942$, $0.984$
$\gamma_{A}$,${\bf 0.867}$, $0.886$, $0.972$, $1.211$, $1.219$
$\gamma_{R}$,${\bf 0.786}$, $0.806$, $0.922$, $1.384$, $1.261$
$\gamma_{C}$,${\bf 0.608}$, $0.644$, $0.899$, $0.787$, $0.935$
$\theta_{A}$,${\bf 0.601}$, $0.627$, $0.907$, $1.088$, $1.119$
$\theta_{R}$,${\bf 0.819}$, $0.833$, $0.931$, $1.175$, $1.122$
$mean$,${\bf 0.585}$, $0.605$, $0.808$, $1.155$, $1.186$
\end{filecontents*}
\csvreader[head to column names, late after line=\\]{accuracy_data_different_models_prior6.csv}{} 
{\csvcoli&\csvcolii&\csvcoliii&\csvcoliv&\csvcolv&\csvcolvi} %
\end{tabular}}
\caption{Mean $E_{\%}$ over the prior range for different ANN architectures for inference based on time series responses of species $\{C\}$, and for ABC parameter inference using summary statistics selected by AS and using all available statistics (in Table \ref{tab:as}). The ABC trial budget mirrors training set size of $3 \times 10^5$ data samples. }
\label{tab:CompArch}
\end{table}


\begin{table}
\small
\centering
\scalebox{1.0}{%
\setlength{\tabcolsep}{7pt}
\begin{tabular}{c|r|c|c}
Architecture & Train Time & \multicolumn{2}{c}{\shortstack{No. of Parameters}} \\
\cline{3-4}
&  & Total & Trainable \\
\hline 
DNN & 26s & 457,167 & 450,767 \\
PEN$_{10}$ & 1m 53s & 385,727 & 383,327   \\
CNN & 4m 41s & 492,415 & 490,015 \\
\end{tabular}}
\caption{Training time and the number of trainable parameters for each architecture for an inference problem based on time series responses of species $\{C\}$. Experiments performed on hardware comprising of 3.6 GHz Intel Core i7 (4 cores) CPU, nVidia GeForce GTX 1080 GPU, 16 GB RAM, running Python3 on Windows 10 operating system.}
\label{tab:CompArch2}
\end{table}


\subsubsection{The effect of the observed species on inference quality for the CNN}
\begin{figure*}
\subfloat[Single mRNAs]{\label{fig:single_mrna}\includegraphics[width=.3\linewidth]{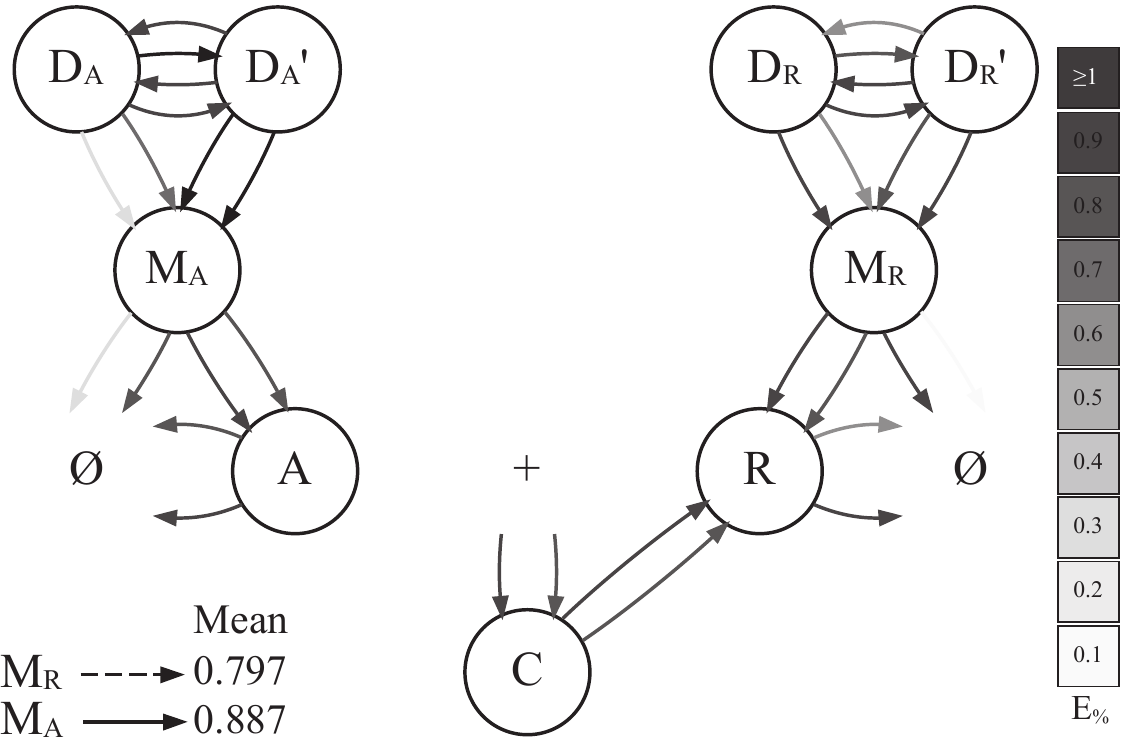}}
\hfill
\subfloat[Single proteins]{\label{fig:single_proteins}\includegraphics[width=.3\linewidth]{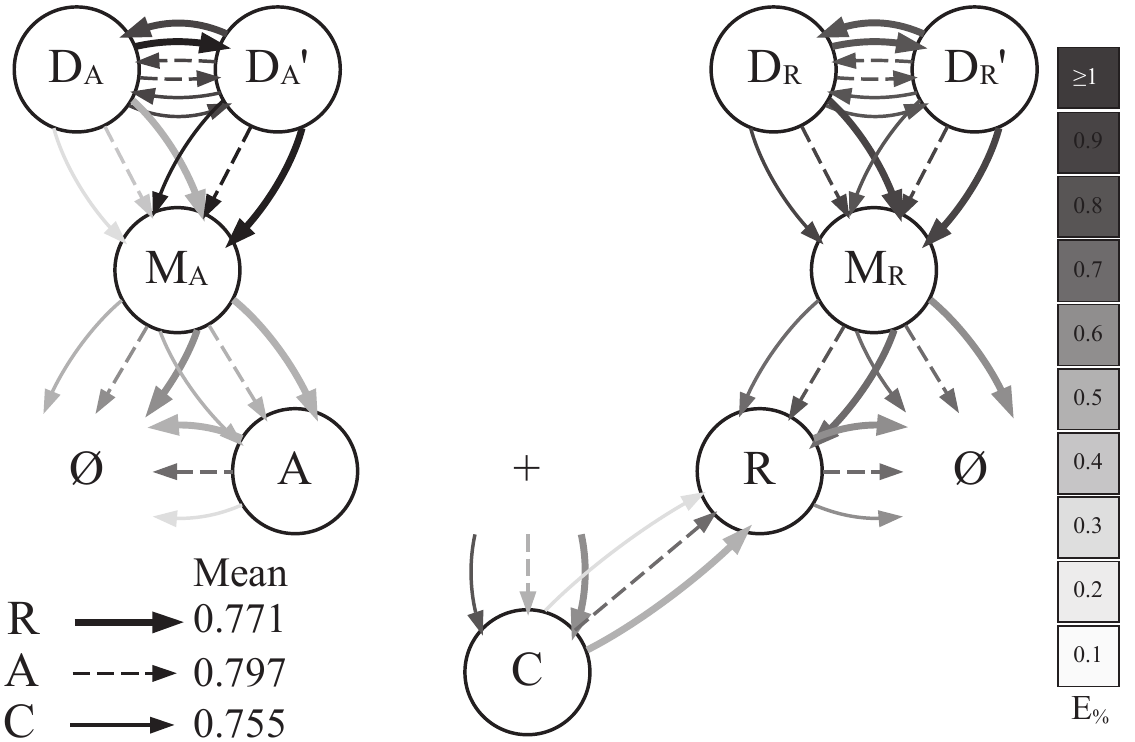}}
\hfill
\subfloat[Some combinations of mRNA and proteins]{\label{fig:comb_mrna_pro}\includegraphics[width=.3\linewidth]{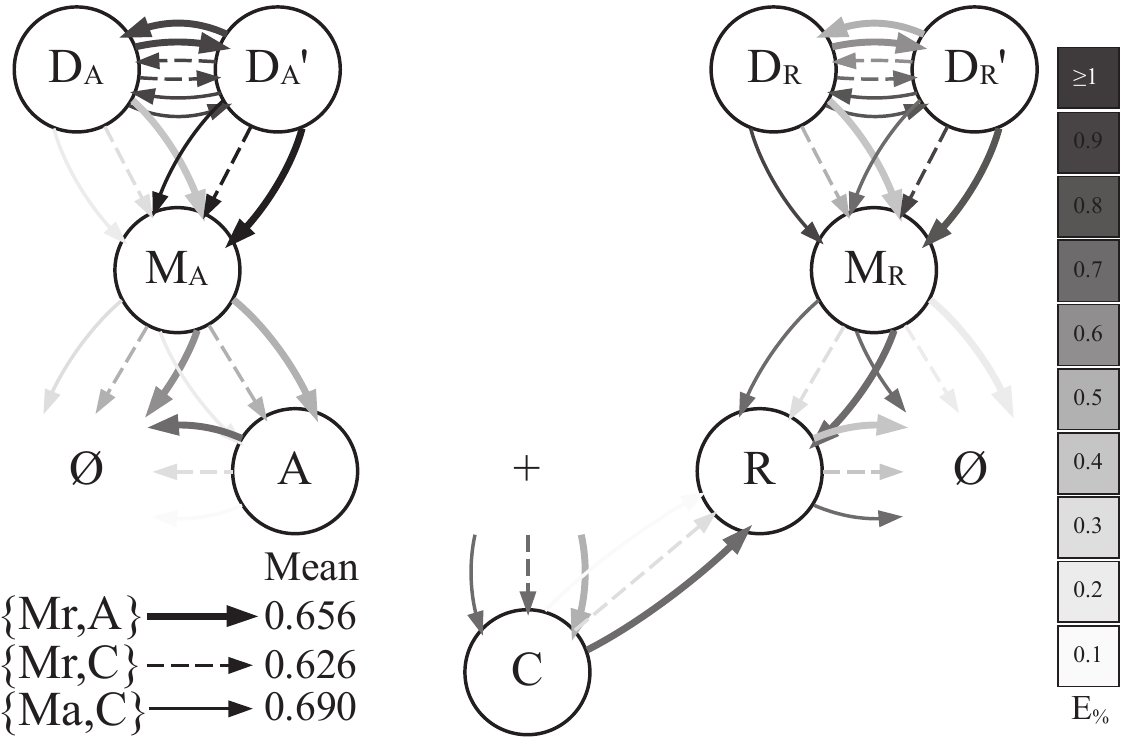}}

\subfloat[Some combinations of proteins]{\label{fig:comb_pro}\includegraphics[width=.3\linewidth]{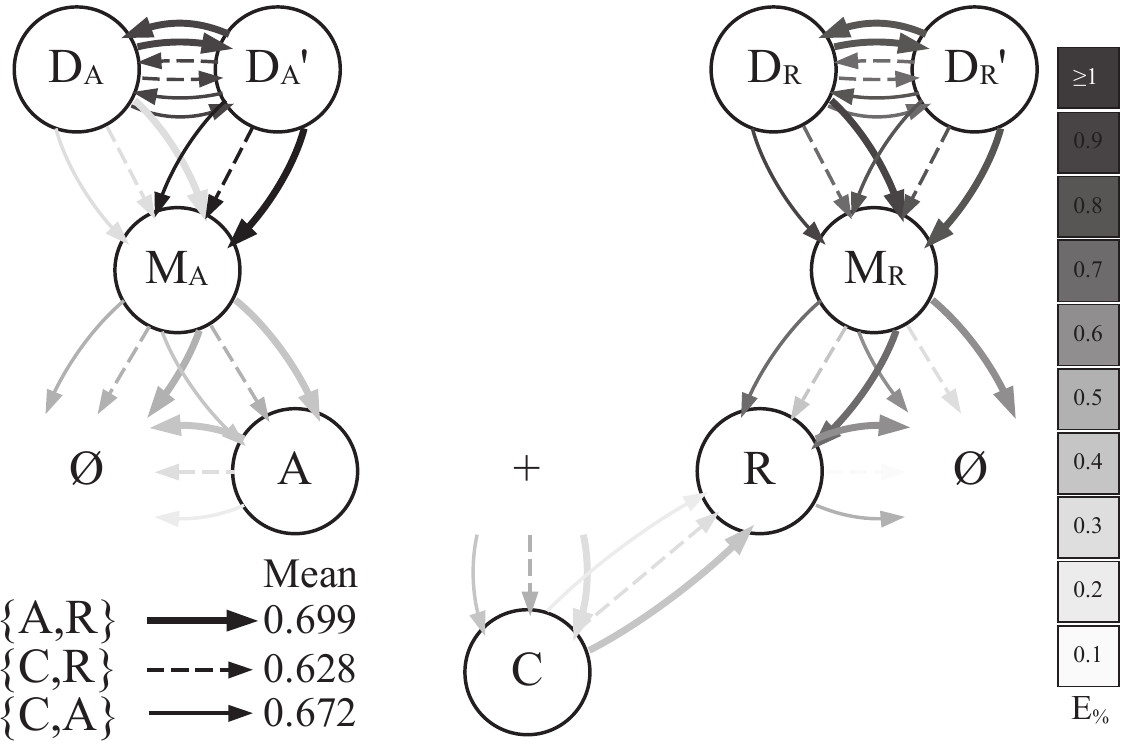}}
\hfill
\subfloat[Some combinations of 3 species]{\label{fig:comb_3}\includegraphics[width=.3\linewidth]{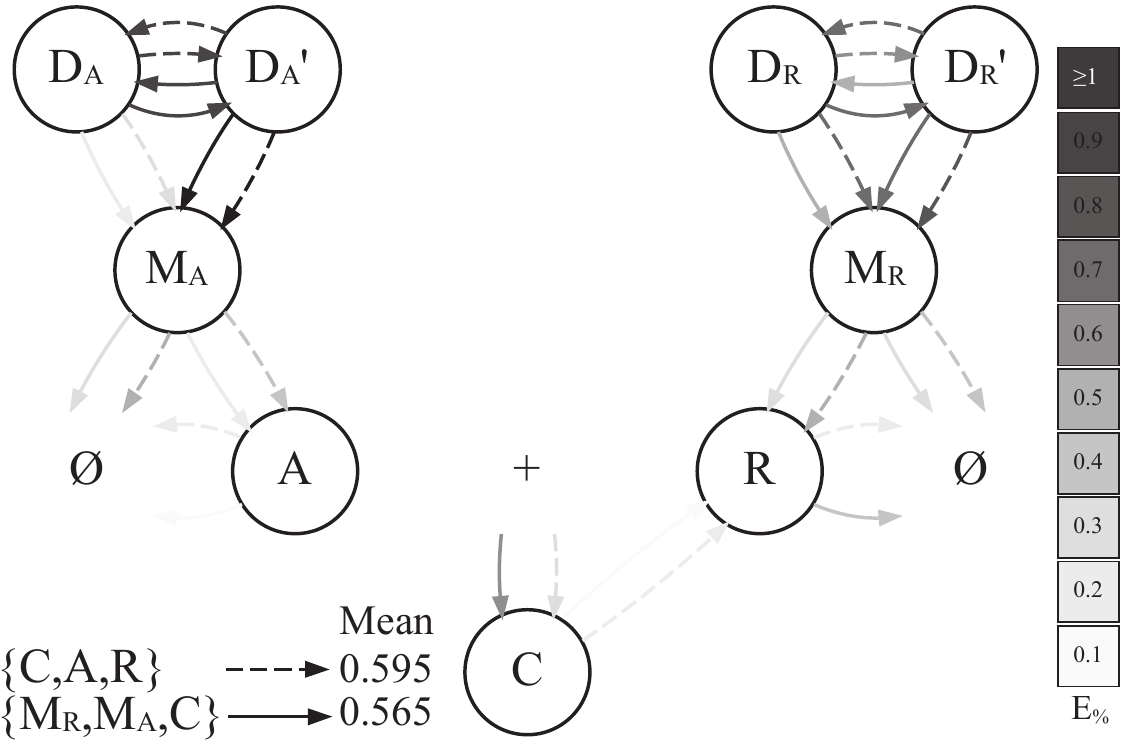}}
\hfill
\subfloat[All mRNA and proteins, different architectures]{\label{fig:comb_5}\includegraphics[width=.3\linewidth]{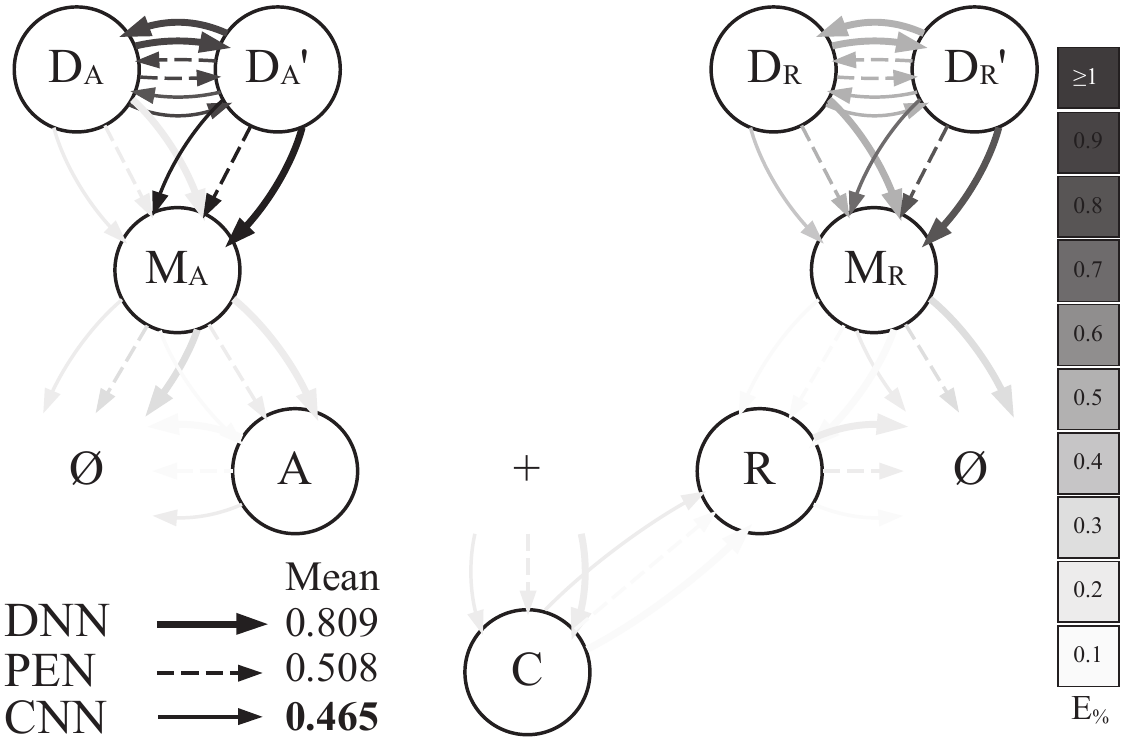}}
\hfill

\caption{Single and multi-species input to CNN. Each edge in the graph corresponds to the particular species used. $E_\%$ values are mapped to a color scheme seen in graph (f), where lighter shades correspond to low values and darker to high values. If $E_\% > 1$ the edge color is black. Graph (f) uses all mRNA and proteins available in the model and compare $E_\%$ between different architectures.}
\label{fig:species_comb}
\end{figure*}

Next we conducted a series of experiments in which we used the CNN and varied larger sets of observed species compared to Table \ref{tab:ts_step} (either single species or combinations of species). The purpose of this was to gain insight into whether or not we can improve inference quality for certain parameters by including certain species or combinations of species as input to the CNN. Setup 1 was used for all 3 architectures for this experiment.
Figure \ref{fig:species_comb} shows a mapping of inference quality in terms of $E_{\%}$ per parameter to the networks edges (see Figure \ref{fig:vilar_net} for reference).
We first looked at the inference quality when observing a single species. Figure \ref{fig:single_mrna} and \ref{fig:single_proteins} lists the posterior estimation error values in terms of $E_{\%}$ corresponding to each mRNA and protein species (single subsets). This entails training one CNN model for those species. 
It can be seen that species $\{C\}$ results in overall least error in estimating the posterior mean, which is not surprising since $\{C\}$ is the final product and common component of the biochemical network. However, certain species are more informative towards inferring certain parameters, which is intuitive considering species-parameter reaction relationships within the genetic oscillator. For example, the rate parameters associated with translation and degradation of proteins get a small increase in quality when single proteins are used as input to the CNN. Similarly, we observe
slightly better performance for rate parameters associated with transcription and degradation of mRNAs when including mRNAs as input. Intuitively, if we combine mRNAs and proteins in combinations of two as demonstrated in figure \ref{fig:comb_mrna_pro} one can expect to get a combined performance from Figure \ref{fig:single_mrna} and \ref{fig:single_proteins}. We observe only a small increase in performance for these combinations. Again, as proteins in combinations of two are used, we observe an increase of performance for rate parameters directly coupled to proteins reactions. As we increase the the size of combinations to 3 (Figure \ref{fig:comb_3}) and 5 (Figure \ref{fig:comb_5}) the results outperform the quality of lower dimensional combinations. This can be motivated by the fact that several species are needed to infer the highly non-linear complexity within the model and we start to see benefits of the expressive power and scalability of CNN for high dimensional problems. In Figure \ref{fig:comb_5} we also compare the performance to other architectures, where the CNN stands as the best performing architecture. The reason $\alpha'_A$ parameter is so difficult to infer (black edges in Figure \ref{fig:species_comb}) is unfortunately out of our comprehension.   

In practice, simultaneously observing the trajectories of more than one species is experimentally challenging but recent advances in single-cell quantification of both RNA and protein levels are promising \cite{lin_ultra-sensitive_2019,Rei_2019,kays_protein_2019}. 


%


\begin{table}
\centering
\scalebox{1}{%
\setlength{\tabcolsep}{5pt}
\begin{tabular}{c|c|c|c|c}%
& \multicolumn{4}{c}{\shortstack{Final step (h)}} \\
\cline{2-5}
Step(h) & 25 & 50 & 100 & 200\\
\cline{1-5}
& \multicolumn{4}{c}{\shortstack{CNN - Species $\{C\}$}} \\
\cline{1-5}
\begin{filecontents*}{cnn_1sp_range_vs_time_stdMeanE_CNN.csv}
,$25$,$50$,$100$,$200$
$0.5$,${\bf 0.831\pm0.004}$,${\bf 0.804\pm0.004}$,${\bf 0.780\pm0.003}$,${\bf 0.762\pm0.005}$
$1.0$,${\bf 0.862\pm0.004}$,${\bf 0.839\pm0.003}$,${\bf 0.816\pm0.003}$,${\bf 0.799\pm0.009}$
$2.0$,${\bf 0.893\pm0.004}$,${\bf 0.877\pm0.004}$,${\bf 0.861\pm0.004}$,${\bf 0.842\pm0.003}$
\end{filecontents*}
\csvreader[head to column names, before reading=\sisetup{round-mode=places,round-precision=3}, late after line=\\]{cnn_1sp_range_vs_time_stdMeanE_CNN.csv}{}%
{\csvcoli&\csvcolii&\csvcoliii&\csvcoliv&\csvcolv}%
& \multicolumn{4}{c}{\shortstack{PEN$_{10}$ - Species $\{C\}$}} \\
\cline{1-5}
\begin{filecontents*}{pen_1sp_range_vs_time_stdMeanE_PEN.csv}
,$25$,$50$,$100$,$200$
$0.5$,$0.842\pm0.004$,$0.821\pm0.007$,$0.810\pm0.011$,$0.801\pm0.022$
$1.0$,$0.868\pm0.003$,$0.848\pm0.005$,$0.834\pm0.006$,$0.818\pm0.007$
$2.0$,$0.897\pm0.003$,$0.882\pm0.004$,$0.870\pm0.004$,$0.862\pm0.009$
\end{filecontents*}
\csvreader[head to column names, before reading=\sisetup{round-mode=places,round-precision=3}, late after line=\\]{pen_1sp_range_vs_time_stdMeanE_PEN.csv}{}%
{\csvcoli&\csvcolii&\csvcoliii&\csvcoliv&\csvcolv}%
& \multicolumn{4}{c}{\shortstack{DNN - Species $\{C\}$}} \\
\cline{1-5}
\begin{filecontents*}{dnn_1sp_range_vs_time_stdMeanE_DNN.csv}
,$25$,$50$,$100$,$200$
$0.5$,$0.882\pm0.007$,$0.904\pm0.004$,$0.922\pm0.005$,$0.945\pm0.014$
$1.0$,$0.886\pm0.002$,$0.900\pm0.006$,$0.927\pm0.008$,$0.944\pm0.009$
$2.0$,$0.906\pm0.003$,$0.911\pm0.003$,$0.931\pm0.004$,$0.948\pm0.007$
\end{filecontents*}
\csvreader[head to column names, before reading=\sisetup{round-mode=places,round-precision=3}, late after line=\\]{dnn_1sp_range_vs_time_stdMeanE_DNN.csv}{}%
{\csvcoli&\csvcolii&\csvcoliii&\csvcoliv&\csvcolv}%

\cline{1-5}
\noalign{\vspace{10pt}}
& \multicolumn{4}{c}{\shortstack{CNN - Species $\{C, A, R\}$}} \\
\cline{1-5}
\begin{filecontents*}{cnn_3sp_range_vs_time_stdMeanE_CNN.csv}
,$25$,$50$,$100$,$200$
$0.5$,${\bf 0.692\pm0.004}$,${\bf 0.654\pm0.007}$,${\bf 0.630\pm0.010}$,${\bf 0.604\pm0.011}$
$1.0$,${\bf 0.728\pm0.005}$,${\bf 0.697\pm0.007}$,${\bf 0.662\pm0.009}$,${\bf 0.636\pm0.014}$
$2.0$,${\bf 0.770\pm0.003}$,${\bf 0.740\pm0.004}$,${\bf 0.711\pm0.004}$,${\bf 0.684\pm0.008}$
\end{filecontents*}
\csvreader[head to column names, before reading=\sisetup{round-mode=places,round-precision=3}, late after line=\\]{cnn_3sp_range_vs_time_stdMeanE_CNN.csv}{}%
{\csvcoli&\csvcolii&\csvcoliii&\csvcoliv&\csvcolv}%
& \multicolumn{4}{c}{\shortstack{PEN$_{10}$ - Species $\{C, A, R\}$}} \\
\cline{1-5}
\begin{filecontents*}{pen_3sp_range_vs_time_stdMeanE_PEN.csv}
,$25$,$50$,$100$,$200$
$0.5$,$0.705\pm0.006$,$0.677\pm0.009$,$0.657\pm0.009$,$0.653\pm0.041$
$1.0$,$0.737\pm0.004$,$0.711\pm0.006$,$0.686\pm0.006$,$0.668\pm0.012$
$2.0$,$0.774\pm0.003$,$0.749\pm0.005$,$0.725\pm0.005$,$0.708\pm0.010$
\end{filecontents*}
\csvreader[head to column names, before reading=\sisetup{round-mode=places,round-precision=3}, late after line=\\]{pen_3sp_range_vs_time_stdMeanE_PEN.csv}{}%
{\csvcoli&\csvcolii&\csvcoliii&\csvcoliv&\csvcolv}%
& \multicolumn{4}{c}{\shortstack{DNN - Species $\{C, A, R\}$}} \\
\cline{1-5}
\begin{filecontents*}{dnn_3sp_range_vs_time_stdMeanE_DNN.csv}
,$25$,$50$,$100$,$200$
$0.5$,$0.831\pm0.006$,$0.847\pm0.006$,$0.883\pm0.015$,$0.919\pm0.014$
$1.0$,$0.851\pm0.007$,$0.864\pm0.011$,$0.883\pm0.009$,$0.919\pm0.010$
$2.0$,$0.870\pm0.006$,$0.879\pm0.009$,$0.907\pm0.012$,$0.934\pm0.012$
\end{filecontents*}
\csvreader[head to column names, before reading=\sisetup{round-mode=places,round-precision=3}, late after line=\\]{dnn_3sp_range_vs_time_stdMeanE_DNN.csv}{}%
{\csvcoli&\csvcolii&\csvcoliii&\csvcoliv&\csvcolv}%

\cline{1-5}
\noalign{\vspace{10pt}}
& \multicolumn{4}{c}{\shortstack{CNN - Species $\{M_A, M_R, C, A, R\}$}} \\
\cline{1-5}
\begin{filecontents*}{cnn_5sp_range_vs_time_stdMeanE_CNN.csv}
,$25$,$50$,$100$,$200$
$0.5$,$0.565\pm0.012$,${\bf 0.530\pm0.014}$,${\bf 0.503\pm0.012}$,${\bf 0.471\pm0.023}$
$1.0$,${\bf 0.605\pm0.005}$,$0.570\pm0.008$,${\bf 0.545\pm0.012}$,${\bf 0.507\pm0.009}$
$2.0$,${\bf 0.648\pm0.003}$,${\bf 0.619\pm0.005}$,${\bf 0.587\pm0.005}$,${\bf 0.557\pm0.009}$
\end{filecontents*}
\csvreader[head to column names, before reading=\sisetup{round-mode=places,round-precision=3}, late after line=\\]{cnn_5sp_range_vs_time_stdMeanE_CNN.csv}{}%
{\csvcoli&\csvcolii&\csvcoliii&\csvcoliv&\csvcolv}%
& \multicolumn{4}{c}{\shortstack{PEN$_{10}$ - Species $\{M_A, M_R, C, A, R\}$}} \\
\cline{1-5}
\begin{filecontents*}{pen_5sp_range_vs_time_stdMeanE_PEN.csv}
,$25$,$50$,$100$,$200$
$0.5$,${\bf 0.563\pm0.004}$,$0.538\pm0.008$,$0.517\pm0.009$,$0.513\pm0.053$
$1.0$,$0.605\pm0.006$,${\bf 0.568\pm0.004}$,$0.547\pm0.010$,$0.527\pm0.012$
$2.0$,$0.659\pm0.004$,$0.623\pm0.006$,$0.593\pm0.007$,$0.566\pm0.007$
\end{filecontents*}
\csvreader[head to column names, before reading=\sisetup{round-mode=places,round-precision=3}, late after line=\\]{pen_5sp_range_vs_time_stdMeanE_PEN.csv}{}%
{\csvcoli&\csvcolii&\csvcoliii&\csvcoliv&\csvcolv}%
& \multicolumn{4}{c}{\shortstack{DNN - Species $\{M_A, M_R, C, A, R\}$}} \\
\cline{1-5}
\begin{filecontents*}{dnn_5sp_range_vs_time_stdMeanE_DNN.csv}
,$25$,$50$,$100$,$200$
$0.5$,$0.685\pm0.009$,$0.713\pm0.008$,$0.756\pm0.018$,$0.803\pm0.019$
$1.0$,$0.715\pm0.008$,$0.731\pm0.007$,$0.777\pm0.020$,$0.806\pm0.021$
$2.0$,$0.752\pm0.010$,$0.761\pm0.008$,$0.785\pm0.011$,$0.823\pm0.010$
\end{filecontents*}
\csvreader[head to column names, before reading=\sisetup{round-mode=places,round-precision=3}, late after line=\\]{dnn_5sp_range_vs_time_stdMeanE_DNN.csv}{}%
{\csvcoli&\csvcolii&\csvcoliii&\csvcoliv&\csvcolv}%
\end{tabular}}
\caption{Mean $E_{\%}$ over 10 different training ($3 \times 10^5$ samples), validation ($2 \times 10^4$) and test ($10^5$ samples) datasets with simulations of varying step sizes (temporal sampling frequency), final simulation termination and species involved.}
\label{tab:ts_step}
\end{table}

\begin{table}
\centering
\scalebox{1}{%
\setlength{\tabcolsep}{5pt}
\begin{tabular}{c|c|c|c|c|c|c}%
& \multicolumn{6}{c}{\shortstack{Final step interval (h)}} \\
\cline{2-7}
Species, Step(h) & 25-50 & 50-100 & 100-200 & 25-50 & 50-100 & 100-200\\
\cline{1-7}
$\{C\}$ & \multicolumn{3}{c|}{\shortstack{CNN}} & \multicolumn{3}{c}{\shortstack{PEN}} \\
\cline{1-7}
\begin{filecontents*}{range_change_all_1sp.csv}
, $25-50$, $50-100$, $100-200$, $25-50$, $50-100$, $100-200$
$0.5$, ${\bf -3.358}$, ${\bf -3.077}$, ${\bf -2.362}$, $-2.558$, $-1.358$, $-1.124$
$1.0$, ${\bf -2.741}$, ${\bf -2.819}$, ${\bf -2.128}$, $-2.358$, $-1.679$, $-1.956$
$2.0$, ${\bf -1.824}$, ${\bf -1.858}$ , ${\bf -2.257}$, $-1.701$, $-1.379$, $-0.928$
\end{filecontents*}
\csvreader[head to column names, before reading=\sisetup{round-mode=places,round-precision=3}, late after line=\\]{range_change_all_1sp.csv}{}%
{\csvcoli&\csvcolii&\csvcoliii&\csvcoliv&\csvcolv&\csvcolvi&\csvcolvii}%

\noalign{\vspace{10pt}}
$\{C, A, R\}$ & \multicolumn{3}{c|}{\shortstack{CNN}} & \multicolumn{3}{c}{\shortstack{PEN}} \\
\cline{1-7}
\begin{filecontents*}{range_change_all_3sp.csv}
, $25-50$, $50-100$, $100-200$, $25-50$, $50-100$, $100-200$
$0.5$, ${\bf -5.810}$, ${\bf -3.810}$, ${\bf -4.305}$, $-4.136$, $-3.044$, $-0.613$
$1.0$, ${\bf -4.448}$, ${\bf -5.287}$, ${\bf -4.088}$, $-3.657$, $-3.644$, $-2.695$
$2.0$, ${\bf -4.054}$, ${\bf -4.079}$, ${\bf -3.947}$, $-3.338$, $-3.310$, $-2.401$
\end{filecontents*}
\csvreader[head to column names, before reading=\sisetup{round-mode=places,round-precision=3}, late after line=\\]{range_change_all_3sp.csv}{}%
{\csvcoli&\csvcolii&\csvcoliii&\csvcoliv&\csvcolv&\csvcolvi&\csvcolvii}%

\noalign{\vspace{10pt}}
$\{M_A, M_R, C, A, R\}$ & \multicolumn{3}{c|}{\shortstack{CNN}} & \multicolumn{3}{c}{\shortstack{PEN}} \\
\cline{1-7}
\begin{filecontents*}{range_change_all_5sp.csv}
, $25-50$, $50-100$, $100-200$, $25-50$, $50-100$, $100-200$
$0.5$, ${\bf -6.604}$, ${\bf -5.368}$, ${\bf -6.794}$, $-4.647$, $-4.061$, $-0.780$
$1.0$, $-6.140$, ${\bf -4.587}$, ${\bf -7.495}$, ${\bf -6.514}$, $-3.839$, $-3.795$
$2.0$, $-4.685$, ${\bf -5.451}$, ${\bf -5.386}$, ${\bf -5.778}$, $-5.059$, $-4.770$
\end{filecontents*}
\csvreader[head to column names, before reading=\sisetup{round-mode=places,round-precision=3}, late after line=\\]{range_change_all_5sp.csv}{}%
{\csvcoli&\csvcolii&\csvcoliii&\csvcoliv&\csvcolv&\csvcolvi&\csvcolvii}%
\end{tabular}}
\caption{Relative percentage change in mean $E_{\%}$ over different final termination steps for results in Table \ref{tab:ts_step}. The change in calculated relative to $a$ for the step from $a$ to $b$.}
\label{tab:ts_percent_change_endstep}
\end{table}

For further discussion around using 5 species (all mRNAs and proteins), we refer the readers attention back to Table \ref{tab:ts_step} (the relationship between simulation resolution in terms of or step size and total simulation time). We observe that as the step size increases and the final step decreases, the inference quality declines. This is intuitive as higher temporal resolution allows the convolution operator to characterize more detailed and accurate features over the input time series. This allows the CNN to incorporate more degrees of differentiation between the fine patterns present within time series from the genetic oscillator, and how they affect parameters $\theta$. Also, the results are intuitive as longer simulation lengths will incorporate distinct oscillating patterns (e.g., see Figure \ref{fig:vilar_ts}) with larger periods, providing better discrimination abilities to the CNN. 
As a final evaluation and in order to gauge the potential gains in mean $E_{\%}$ using setup 2, CNN models for 3 (\{C,A,R\}, practical in scenarios where protein levels are measurable) and 5 (\{Ma, Mr, C, A, R\}, our best performing combination) species were evaluated. The resulting mean $E_{\%}$ values are 0.535 and 0.405 respectively. The use of setup 2 delivered these gains at the cost of $\sim 3.5$ times increase in training time.

\subsubsection{Effect of training data size on inference quality}
We next focus on the actual cost of performing inference with the ANNs. Since the main cost is to  generate the simulated training data, we study the effect of the training set size on the inference performance.

\begin{table}
\centering
\scalebox{1.0}{%
\setlength{\tabcolsep}{8pt}
\begin{tabular}{c|c|c|c|c}%
    Param. & \multicolumn{4}{c}{\shortstack{No. of Training Samples}} \\
    \cline{2-5}

    & {30k} & {100k }& {200k} & {300k} 
    \\\hline
    \begin{filecontents*}{accuracy_data_different_training_sizes_prior6.csv}
0,1,2,3,4,5,6,7,8
$\alpha_{A}$,9.119,0.456,8.213,0.411,7.897,0.395,$\mathbf{7.832}$,$\mathbf{0.392}$
$\alpha_{A}'$,77.965,0.624,66.956,0.536,65.181,0.521,$\mathbf{63.971}$,$\mathbf{0.512}$
$\alpha_{R}$,0.975,0.975,0.937,0.937,0.925,0.925,$\mathbf{0.920}$,$\mathbf{0.920}$
$\alpha_{R}'$,8.835,0.883,7.782,0.778,7.627,0.763,$\mathbf{7.583}$,$\mathbf{0.758}$
$\beta_{A}$,7.506,0.601,6.649,0.532,6.411,0.513,$\mathbf{6.315}$,$\mathbf{0.505}$
$\beta_{R}$,0.899,0.600,0.764,0.509,0.737,0.491,$\mathbf{0.735}$,$\mathbf{0.490}$
$\delta_{MA}$,1.556,0.566,1.410,0.513,1.369,0.498,$\mathbf{1.359}$,$\mathbf{0.494}$
$\delta_{MR}$,0.252,0.503,0.213,0.427,$\mathbf{0.201}$,$\mathbf{0.403}$,0.201,0.403
$\delta_{A}$,0.241,0.321,0.205,0.274,0.196,0.261,$\mathbf{0.191}$,$\mathbf{0.255}$
$\delta_{R}$,0.082,0.466,0.069,0.397,0.066,0.377,$\mathbf{0.064}$,$\mathbf{0.366}$
$\gamma_{A}$,0.471,0.942,0.442,0.884,0.436,0.872,$\mathbf{0.434}$,$\mathbf{0.867}$
$\gamma_{R}$,0.867,0.867,0.802,0.802,0.790,0.790,$\mathbf{0.786}$,$\mathbf{0.786}$
$\gamma_{C}$,0.545,0.726,0.485,0.646,0.466,0.621,$\mathbf{0.456}$,$\mathbf{0.608}$
$\theta_{A}$,12.217,0.698,10.921,0.624,10.614,0.606,$\mathbf{10.525}$,$\mathbf{0.601}$
$\theta_{R}$,67.207,0.896,62.933,0.839,61.963,0.826,$\mathbf{61.406}$,$\mathbf{0.819}$
mean,12.582,0.675,11.252,0.607,10.992,0.591,$\mathbf{10.852}$,$\mathbf{0.585}$
\end{filecontents*}
\csvreader[head to column names, late after line=\\]{accuracy_data_different_training_sizes_prior6.csv}{}
    {\csvcoli&\csvcoliii&\csvcolv&\csvcolvii&\csvcolix} 

    \end{tabular}}
    \caption{$E_{\%}$ on the test set over the prior range for different sizes of training data for the inference task based on observing species $\{C\}$ using the CNN architecture.}
    \label{tab:ds_sizes}
    \end{table}

Table \ref{tab:ds_sizes} depicts the relationship between the size of the training set and inference error in the form of $E_{\%}$. The CNN architecture trained using setup 2 is used to study the effect of varying training set sizes. The largest training set size ($3 \times 10^5$ samples) leads to the least error, but the improvements over a training set of $2 \times 10^5$ samples are negligible. The most significant step up in inference accuracy is reflected when moving from a training set of $3 \times 10^4$ samples to $10^5$ samples. For the considered problem, the training set size of $10^5$ samples appears to strike a fine balance between error in estimating the posterior mean and required training set size.

\section{Conclusion}
\label{sec:conclusion}
This paper presented the convolutional neural networks architecture for learning summary statistics for use in approximate Bayesian computation. In general, the proposed summary statistic learning framework can be used in any likelihood-free parameter inference framework that makes use of summary statistics to compare observed data and simulated responses. The network learns the mapping from time series responses ${\bf y} = f(\theta)$ to control parameters $\theta$, which characterizes the estimated posterior mean and effectively represents the learned summary statistics. The proposed convolutional architecture is compared to state-of-the-art deep neural network and partially exchangeable network architectures on two small-scale benchmark test problems and a large-scale high-dimensional biochemical reaction network example. All three architectures perform well on small-scale test problems (the moving averages MA(2) and Lotka-Volterra predator-prey models), while the proposed convolutional architecture outperforms existing approaches in case of the large-scale high-dimensional stochastic biochemical reaction network test problem. The proposed architecture is shown to be robust and versatile with respect to varying problem complexity and training set size. In systems biology the parameter inference problem is highly interesting with the rapid improvements in high-throughput experimental techniques to observe single-cell, temporal and molecular-level data. However, there is a lack of benchmark problems of sufficient complexity. In this paper we have empirically and systematically assessed inference quality for a complex high-dimensional network \cite{vilar2002mechanisms} under various assumptions on the data quality and richness. It is our hope that this will also serve general methods development in systems biology well by providing a documented benchmark with real-world relevance. As future work, we plan to further develop ANN model-driven approaches and to introduce adaptive sampling algorithms for obtaining efficient training data for the model, and automatic hyperparameter optimization of the ANN models.



%



\section*{Acknowledgment}
The work was funded by the NIH under grant no. NIH/2R01EB014877-04A1, the eSSENCE strategic collaboration on eScience, and the G{\"o}ran Gustafsson foundation.

\newpage
\bibliographystyle{plainnat}
\bibliography{bibfile}

\begin{thebibliography}{29}
\providecommand{\natexlab}[1]{#1}
\providecommand{\url}[1]{\texttt{#1}}
\expandafter\ifx\csname urlstyle\endcsname\relax
  \providecommand{\doi}[1]{doi: #1}\else
  \providecommand{\doi}{doi: \begingroup \urlstyle{rm}\Url}\fi

\bibitem[Beaumont et~al.(2002)Beaumont, Zhang, and
  Balding]{beaumont2002approximate}
Mark~A Beaumont, Wenyang Zhang, and David~J Balding.
\newblock Approximate bayesian computation in population genetics.
\newblock \emph{Genetics}, 162\penalty0 (4):\penalty0 2025--2035, 2002.

\bibitem[Cornuet et~al.(2008)Cornuet, Santos, Beaumont, Robert, Marin, Balding,
  Guillemaud, and Estoup]{cornuet2008inferring}
Jean-Marie Cornuet, Filipe Santos, Mark~A Beaumont, Christian~P Robert,
  Jean-Michel Marin, David~J Balding, Thomas Guillemaud, and Arnaud Estoup.
\newblock Inferring population history with diy abc: a user-friendly approach
  to approximate bayesian computation.
\newblock \emph{Bioinformatics}, 24\penalty0 (23):\penalty0 2713--2719, 2008.

\bibitem[Cranmer et~al.(2020{\natexlab{a}})Cranmer, Brehmer, and
  Louppe]{Cranmer2020}
Kyle Cranmer, Johann Brehmer, and Gilles Louppe.
\newblock The frontier of simulation-based inference.
\newblock \emph{Proceedings of the National Academy of Sciences},
  2020{\natexlab{a}}.
\newblock ISSN 0027-8424.
\newblock \doi{10.1073/pnas.1912789117}.

\bibitem[Cranmer et~al.(2020{\natexlab{b}})Cranmer, Brehmer, and
  Louppe]{Cranmer_frontierSBI}
Kyle Cranmer, Johann Brehmer, and Gilles Louppe.
\newblock The frontier of simulation-based inference.
\newblock \emph{Proceedings of the National Academy of Sciences}, 117\penalty0
  (48):\penalty0 30055--30062, 2020{\natexlab{b}}.
\newblock ISSN 0027-8424.
\newblock \doi{10.1073/pnas.1912789117}.
\newblock URL \url{https://www.pnas.org/content/117/48/30055}.

\bibitem[Drawert et~al.(2016)Drawert, Hellander, Bales, Banerjee, Bellesia,
  Daigle~Jr, Douglas, Gu, Gupta, Hellander, et~al.]{drawert2016stochastic}
Brian Drawert, Andreas Hellander, Ben Bales, Debjani Banerjee, Giovanni
  Bellesia, Bernie~J Daigle~Jr, Geoffrey Douglas, Mengyuan Gu, Anand Gupta,
  Stefan Hellander, et~al.
\newblock Stochastic simulation service: bridging the gap between the
  computational expert and the biologist.
\newblock \emph{PLoS computational biology}, 12\penalty0 (12):\penalty0
  e1005220, 2016.

\bibitem[Fearnhead and Prangle(2012)]{fearnhead2012constructing}
Paul Fearnhead and Dennis Prangle.
\newblock Constructing summary statistics for approximate bayesian computation:
  semi-automatic approximate bayesian computation.
\newblock \emph{Journal of the Royal Statistical Society: Series B (Statistical
  Methodology)}, 74\penalty0 (3):\penalty0 419--474, 2012.

\bibitem[Gillespie(1977)]{gillespie1977exact}
Daniel~T Gillespie.
\newblock Exact stochastic simulation of coupled chemical reactions.
\newblock \emph{The journal of physical chemistry}, 81\penalty0 (25):\penalty0
  2340--2361, 1977.

\bibitem[Goh and Barab{\'a}si(2008)]{goh2008burstiness}
K-I Goh and A-L Barab{\'a}si.
\newblock Burstiness and memory in complex systems.
\newblock \emph{EPL (Europhysics Letters)}, 81\penalty0 (4):\penalty0 48002,
  2008.

\bibitem[Goodfellow et~al.(2016)Goodfellow, Bengio, and
  Courville]{goodfellow2016deep}
Ian Goodfellow, Yoshua Bengio, and Aaron Courville.
\newblock \emph{Deep learning}.
\newblock MIT press, 2016.

\bibitem[Jastrzebski et~al.(2017)Jastrzebski, Kenton, Arpit, Ballas, Fischer,
  Bengio, and Storkey]{jastrzkebski2017three}
Stanislaw Jastrzebski, Zachary Kenton, Devansh Arpit, Nicolas Ballas, Asja
  Fischer, Yoshua Bengio, and Amos Storkey.
\newblock Three factors influencing minima in sgd.
\newblock \emph{arXiv preprint arXiv:1711.04623}, 2017.

\bibitem[Jiang et~al.(2017)Jiang, Wu, Zheng, and Wong]{10.2307/26384090}
Bai Jiang, Tung-Yu Wu, Charles Zheng, and Wing~H. Wong.
\newblock Learning summary statistic for approximate bayesian computation via
  deep neural network.
\newblock \emph{Statistica Sinica}, 27\penalty0 (4):\penalty0 1595--1618, 2017.
\newblock ISSN 10170405, 19968507.

\bibitem[Jiang et~al.(2021)Jiang, Jacob, Geiger, Matthew, Rumsey, Singh, Wrede,
  Yi, Drawert, Hellander, and Petzold]{10.1093/bioinformatics/btab061}
Richard Jiang, Bruno Jacob, Matthew Geiger, Sean Matthew, Bryan Rumsey,
  Prashant Singh, Fredrik Wrede, Tau-Mu Yi, Brian Drawert, Andreas Hellander,
  and Linda Petzold.
\newblock {Epidemiological modeling in StochSS Live!}
\newblock \emph{Bioinformatics}, 01 2021.
\newblock ISSN 1367-4803.
\newblock \doi{10.1093/bioinformatics/btab061}.
\newblock URL \url{https://doi.org/10.1093/bioinformatics/btab061}.
\newblock btab061.

\bibitem[Joyce and Marjoram(2008)]{joyce2008approximately}
Paul Joyce and Paul Marjoram.
\newblock Approximately sufficient statistics and bayesian computation.
\newblock \emph{Statistical applications in genetics and molecular biology},
  7\penalty0 (1), 2008.

\bibitem[Kays and Chen()]{kays_protein_2019}
Ibrahim Kays and Brian~Edwin Chen.
\newblock Protein and {RNA} quantification of multiple genes in single cells.
\newblock 66\penalty0 (1):\penalty0 15--21.
\newblock ISSN 1940-9818.
\newblock \doi{10.2144/btn-2018-0130}.

\bibitem[Lin et~al.()Lin, Jordi, Son, Van~Phan, Drayman, Abasiyanik, Vistain,
  Tu, and Tay]{lin_ultra-sensitive_2019}
Jing Lin, Christian Jordi, Minjun Son, Hoang Van~Phan, Nir Drayman,
  Mustafa~Fatih Abasiyanik, Luke Vistain, Hsiung-Lin Tu, and SavaÅŸ Tay.
\newblock Ultra-sensitive digital quantification of proteins and {mRNA} in
  single cells.
\newblock 10\penalty0 (1):\penalty0 3544.
\newblock ISSN 2041-1723.
\newblock Number: 1 Publisher: Nature Publishing Group.

\bibitem[Lueckmann et~al.(2021)Lueckmann, Boelts, Greenberg, Gon{\c{c}}alves,
  and Macke]{lueckmann2021benchmarking}
Jan-Matthis Lueckmann, Jan Boelts, David~S Greenberg, Pedro~J Gon{\c{c}}alves,
  and Jakob~H Macke.
\newblock Benchmarking simulation-based inference.
\newblock \emph{arXiv preprint arXiv:2101.04653}, 2021.

\bibitem[Marin et~al.(2012)Marin, Pudlo, Robert, and
  Ryder]{marin2012approximate}
Jean-Michel Marin, Pierre Pudlo, Christian~P Robert, and Robin~J Ryder.
\newblock Approximate bayesian computational methods.
\newblock \emph{Statistics and Computing}, 22\penalty0 (6):\penalty0
  1167--1180, 2012.

\bibitem[Park et~al.(2016)Park, Jitkrittum, and Sejdinovic]{park2016k2}
Mijung Park, Wittawat Jitkrittum, and Dino Sejdinovic.
\newblock K2-abc: Approximate bayesian computation with kernel embeddings.
\newblock In \emph{Artificial Intelligence and Statistics}, pages 398--407,
  2016.

\bibitem[Prangle(2015)]{prangle2015summary}
Dennis Prangle.
\newblock Summary statistics in approximate bayesian computation.
\newblock \emph{arXiv preprint arXiv:1512.05633}, 2015.

\bibitem[Prangle et~al.(2017)]{prangle2017adapting}
Dennis Prangle et~al.
\newblock Adapting the abc distance function.
\newblock \emph{Bayesian Analysis}, 12\penalty0 (1):\penalty0 289--309, 2017.

\bibitem[Radev et~al.(2020)Radev, Mertens, Voss, Ardizzone, and
  K{\"o}the]{radev2020bayesflow}
Stefan~T Radev, Ulf~K Mertens, Andreas Voss, Lynton Ardizzone, and Ullrich
  K{\"o}the.
\newblock Bayesflow: Learning complex stochastic models with invertible neural
  networks.
\newblock \emph{IEEE Transactions on Neural Networks and Learning Systems},
  2020.

\bibitem[Reimeg{\r a}rd et~al.(2019)Reimeg{\r a}rd, Danielsson, Tarbier,
  Schuster, Baskaran, Panagiotou, Dahl, Friedl{\"a}nder, and Gallant]{Rei_2019}
Johan Reimeg{\r a}rd, Marcus Danielsson, Marcel Tarbier, Jens Schuster,
  Sathishkumar Baskaran, Styliani Panagiotou, Niklas Dahl, Marc~R.
  Friedl{\"a}nder, and Caroline~J. Gallant.
\newblock Combined mrna and protein single cell analysis in a dynamic cellular
  system using sparc.
\newblock \emph{bioRxiv}, 2019.
\newblock \doi{10.1101/749473}.

\bibitem[Singh et~al.(2020)Singh, Wrede, and Hellander]{sciope}
Prashant Singh, Fredrik Wrede, and Andreas Hellander.
\newblock {Scalable machine learning-assisted model exploration and inference
  using Sciope}.
\newblock \emph{Bioinformatics}, 07 2020.
\newblock ISSN 1367-4803.

\bibitem[Sisson et~al.(2018)Sisson, Fan, and Beaumont]{sisson2018handbook}
Scott~A Sisson, Yanan Fan, and Mark Beaumont.
\newblock \emph{Handbook of approximate Bayesian computation}.
\newblock Chapman and Hall/CRC, 2018.

\bibitem[Vilar et~al.(2002)Vilar, Kueh, Barkai, and
  Leibler]{vilar2002mechanisms}
Jos{\'e}~MG Vilar, Hao~Yuan Kueh, Naama Barkai, and Stanislas Leibler.
\newblock Mechanisms of noise-resistance in genetic oscillators.
\newblock \emph{Proceedings of the National Academy of Sciences}, 99\penalty0
  (9):\penalty0 5988--5992, 2002.

\bibitem[Wiqvist et~al.(2019)Wiqvist, Mattei, Picchini, and
  Frellsen]{2019arXiv190110230W}
Samuel Wiqvist, Pierre-Alexandre Mattei, Umberto Picchini, and Jes Frellsen.
\newblock Partially exchangeable networks and architectures for learning
  summary statistics in approximate bayesian computation.
\newblock In \emph{International Conference on Machine Learning}, pages
  6798--6807, 2019.

\bibitem[Wrede and Hellander(2019)]{Wredesmart2019}
Fredrik Wrede and Andreas Hellander.
\newblock {Smart computational exploration of stochastic gene regulatory
  network models using human-in-the-loop semi-supervised learning}.
\newblock \emph{Bioinformatics}, 35\penalty0 (24):\penalty0 5199--5206, 05
  2019.
\newblock ISSN 1367-4803.

\bibitem[Xing et~al.(2018)Xing, Arpit, Tsirigotis, and Bengio]{xing2018walk}
Chen Xing, Devansh Arpit, Christos Tsirigotis, and Yoshua Bengio.
\newblock A walk with sgd.
\newblock \emph{arXiv preprint arXiv:1802.08770}, 2018.

\bibitem[Zhou and Chellappa(1988)]{zhou1988computation}
Yi-Tong Zhou and Rama Chellappa.
\newblock Computation of optical flow using a neural network.
\newblock In \emph{IEEE International Conference on Neural Networks}, volume
  1998, pages 71--78, 1988.

\end{thebibliography}


\end{document}